\documentclass[11pt]{article}
\usepackage[utf8]{inputenc}
\usepackage[utf8]{inputenc}

\usepackage[preprint]{acl}
\usepackage{times}
\usepackage{latexsym}
\usepackage[T1]{fontenc}
\usepackage[utf8]{inputenc}
\usepackage{microtype}
\usepackage{inconsolata}
\usepackage{graphicx}
\usepackage{svg}
\usepackage{amsmath}
\usepackage{amssymb}
\usepackage{booktabs}
\usepackage{multirow}
\usepackage{algorithm}
\usepackage{algpseudocode}
\usepackage{xspace}
\usepackage[most]{tcolorbox}
\usepackage[table]{xcolor}
\usepackage{xcolor}
\usepackage{booktabs}
\usepackage{multirow}
\usepackage{titletoc}
\usepackage{array}
\usepackage{listings}
\usepackage{enumitem}
\usepackage{pifont}
\newcommand{\cmark}{\ding{51}}
\newcommand{\xmark}{\ding{55}}

\newcommand{\dn}[1]{\textcolor{red}{\scriptsize{$\downarrow$#1}}}
\definecolor{tableblue}{RGB}{70,130,180}
\definecolor{lightblue}{RGB}{235,245,255}
\definecolor{lighterblue}{RGB}{248,251,255}
\definecolor{groupblue}{RGB}{218,235,252}

\newcommand{\qwenlogo}{\raisebox{-0.15em}{\includegraphics[height=1.1em]{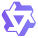}}}
\newcommand{\glmlogo}{\raisebox{-0.15em}{\includegraphics[height=1.1em]{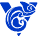}}}
\newcommand{\internvllogo}{\raisebox{-0.15em}{\includegraphics[height=1.1em]{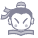}}}

\definecolor{caseblue}{RGB}{232,242,252}
\definecolor{casegreen}{RGB}{226,245,232}
\definecolor{casered}{RGB}{252,228,228}
\definecolor{casegray}{RGB}{245,245,245}

\newcommand{\caseheader}[6]{%
\multicolumn{3}{@{}p{\textwidth}@{}}{%
\begin{minipage}[t]{0.13\textwidth}
\vspace{0pt}
\includegraphics[width=\linewidth]{#1}
\end{minipage}
\hspace{0.018\textwidth}
\begin{minipage}[t]{0.82\textwidth}
\vspace{0pt}
\textbf{#2}
\hfill
\textit{#3}

\vspace{0.35em}
\textbf{Question:} #4

\vspace{0.2em}
\textbf{Options:} #5

\vspace{0.2em}
\textbf{Ground Truth:} #6
\end{minipage}
}%
}

\newcommand{\agentout}[3]{%
\textbf{#1:} \textbf{#2}\par
\textit{#3}\par
\vspace{0.35em}
}

\newcommand{\caseDiagnosis}[1]{%
\vspace{0.25em}
\textbf{Diagnosis:} \textit{#1}
}

\newcommand{\caseFinal}[2]{%
\textbf{Final Answer:} \textbf{#1}\par
\textit{#2}
}

\newcommand{\caseFailure}[2]{%
\textbf{Failure:} \textbf{#1}\par
\textit{#2}
}

\newcommand{\method}{{EAGLE}}

\title{Seeing Before Agreeing: Aligning Multi-Agent Consensus \\ with Visual Evidence}

\author{
\textbf{Yuhan Wang}\textsuperscript{1},
\textbf{Shuochen Chang}\textsuperscript{2},
\textbf{Yalin Feng}\textsuperscript{3},\\
\textbf{Dongsheng Ma}\textsuperscript{1},
\textbf{Yuanzi Li}\textsuperscript{4},
\textbf{Zhengren Wang}\textsuperscript{1},
\textbf{Yinglong Yang}\textsuperscript{5},\\
\textbf{Yufei Chen}\textsuperscript{5},
\textbf{Yikang Wang}\textsuperscript{5},
\textbf{Shaoxu Sun}\textsuperscript{5},
\textbf{Wentao Zhang}\textsuperscript{*\textbf{1}}\\
\textsuperscript{1}Peking University,
\textsuperscript{2}Shanghai Jiao Tong University,\\
\textsuperscript{3}Nanyang Technological University,
\textsuperscript{4}Renmin University of China,
\textsuperscript{5}Shandong University\\
\texttt{fzzh040114@gmail.com}, 
\texttt{wentao.zhang@pku.edu.cn}
}

\begin{document}
\maketitle
\renewcommand{\thefootnote}{\fnsymbol{footnote}}
\footnotetext{* Corresponding author. }
\renewcommand{\thefootnote}{\arabic{footnote}}

\begin{abstract}

Vision-language models (VLMs) have achieved strong performance on visual question answering (VQA). To mitigate individual hallucinations and blind spots, aggregating diverse perspectives via multi-agent collaboration has emerged as a promising paradigm. While this approach has shown great success in textual QA, its potential in the multimodal domain remains under-explored. Existing multi-agent VQA methods predominantly adapt text-centric protocols, focusing on textual discussions while ignoring the alignment of visual information. 
In this work, we reveal a key insight: 
answer-level agreement is insufficient for reliable multi-agent VQA; \textit{aligned visual evidence}---shared support from the image regions agents rely on---is essential for trustworthy consensus.
To leverage this insight, we propose \method{} (\textbf{E}vidence-\textbf{A}ligned \textbf{G}rounded mu\textbf{L}ti-agent r\textbf{E}asoning), a training-free evidence-centered framework for coordinating multiple VLM agents. \method{}
explicitly exposes each agent's grounding regions as visual evidence, enables mutual verification over the evidence, and uses evidence consistency to guide final decision-making.
Experiments on six VQA benchmarks show that \method{} achieves best average performance
across domains
while remaining lightweight,  interpretable, and practical for deployment.
\end{abstract}

\section{Introduction}

Vision-language models (VLMs) have achieved remarkable performance on visual question answering (VQA), demonstrating unprecedented capabilities in complex visual reasoning, chart parsing~\citep{cui2025paddleocr,niu2025mineru2}, and high-resolution scene understanding~\citep{team2025kimi,hong2025glm,wang2025internvl3,bai2025qwen3,an2025llava}. Despite these advancements, VLMs remain prone to hallucinating visual content or relying on spurious language priors and dataset biases when their predictions are not sufficiently grounded in the image context~\citep{goyal2017making, agrawal2018don, li2023evaluating, zhong2024investigating}. Such failures not only lead to unfaithful rationales and limited interpretability~\citep{long2025understanding}, but also cause critical, uncalibrated errors. 
In high-stakes applications of VQA—such as medical image analysis, financial analysis, or legal and document intelligence auditing—such ungrounded reasoning and deceptive consensus are entirely unacceptable, as they can lead to severe and catastrophic consequences~\citep{zhang2025citalaw,lu2025med, gan2025mme,  ma2026citevqa,wang2026agenticocr}. 
Although confidence calibration has been explored to improve the reliability of individual models~\citep{eisenschlos2024selectively, xuan2025seeing, Wang2026EvaluatingAC}, it does not directly verify whether predictions are grounded in the correct visual evidence, and can still be constrained by the inherent biases of a single model.

\begin{figure*}[t]
    \centering
    \includegraphics[width=\linewidth]{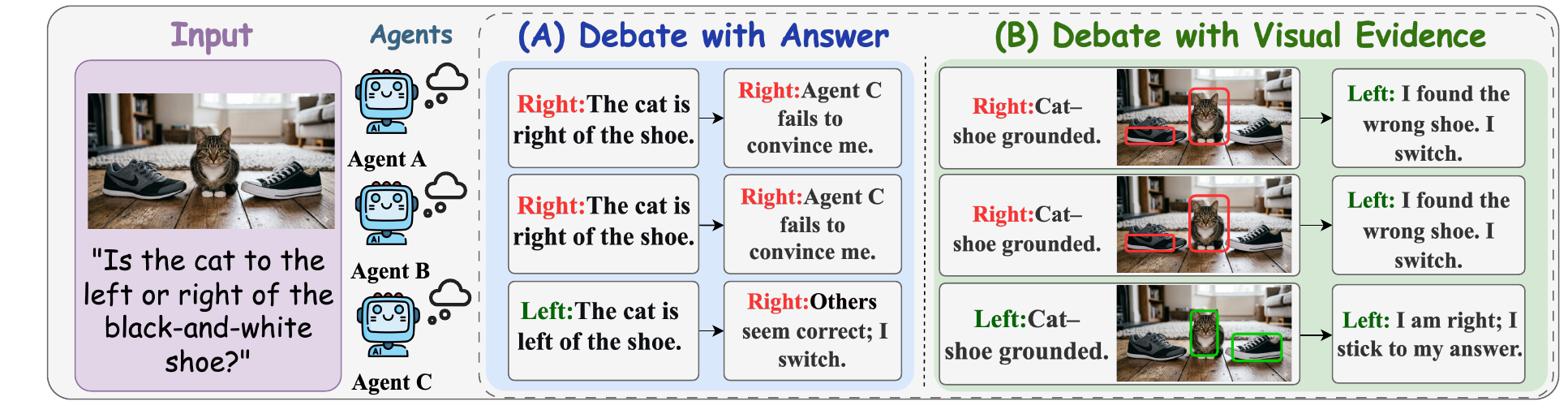}
    \caption{
    \textbf{A case illustrating why answer-level agreement can be misleading.}
    (A) Agents may accept the same textual rationale without verifying whether it is supported by the correct visual regions.
    (B) Explicit grounding makes the supporting evidence comparable, allowing agents to verify whether their agreement is visually aligned.
    }
    \label{fig:motivation}
\end{figure*}

In textual question answering, multi-agent reasoning has emerged as a powerful paradigm to mitigate individual model hallucinations and enhance reliability by aggregating diverse insights through collaborative debate ~\citep{ du2024improving, liang2024encouraging, chen2024reconcile, kaesberg2025voting, pitre2025consensagent, tran2025multi, jiang2026global}. 
Naturally, the community has sought to extend this multi-agent collaborative synergy to VQA tasks, where different models can offer complementary perceptual capabilities—such as identifying distinct objects, spatial relations, or scene-level cues~\citep{rajput2025rethinking, wang2026towards, pandey2026refine}. 
To leverage this complementarity, recent pioneers have introduced specialized agent roles, judge-based aggregation, or external tool orchestration into multi-agent VQA~\citep{jiang2024multi, adhikari2025debating, sivakumaran2026dart}. 
However, these explorations still predominantly inherit the text-centric interaction mechanisms from traditional textual QA. 
They primarily format multi-agent communication as an exchange of language-level rationales, leaving the critical, underlying visual evidence completely implicit and unverified (See Figure~\ref{fig:motivation}(A)).

To uncover the fundamental distinction between VQA and traditional textual tasks within multi-agent contexts, we conduct a pilot study in Section~\ref{sec:pilot}.
Our quantitative analysis reveals a critical, underlying reality: 
reliable multi-agent VQA requires more than answer-level agreement; it also requires \textit{aligned visual evidence}---shared support from the specific image regions that agents rely on---to form trustworthy consensus.
Without explicit anchoring, agents are prone to falling into unstable debates or superficial consensus, as language-level arguments can easily mask divergent visual understandings. 
This motivates our central principle: \textbf{\textit{agents must see and align before agreeing}}---that is, multi-agent consensus should be grounded in aligned visual evidence rather than pure linguistic convergence (see Figure~\ref{fig:motivation}(B)).

Motivated by these observations, we propose \method{} (\textbf{E}vidence-\textbf{A}ligned \textbf{G}rounded mu\textbf{L}ti-agent r\textbf{E}asoning), a universal and training-free multi-agent VQA framework. Fully centered around visual evidence, our pipeline coordinates multiple VLM agents through five core stages: 
\textbf{(1) Evidence Routing} dynamically assigns the appropriate grounding granularity for each question; 
\textbf{(2) Grounded Answer} mandates each agent to output an answer accompanied by grounding boxes and an explicit visual claim explaining how the image regions support the answer; 
\textbf{(3) Evidence Diagnosis} jointly evaluates the consistency of both language responses and visual evidence, allowing for an early exit upon alignment; 
\textbf{(4) Grounded Revision} guides divergent agents to self-correct by cross-verifying their peers' visual evidence and diagnostic signals; and 
\textbf{(5) Evidence-Guided Arbitration} resolves final disagreements by prioritizing the answer backed by the most visually consistent agent group. 
By anchoring multi-agent consensus in explicit visual regions, \method{} delivers highly reliable and interpretable coordinated reasoning.

We evaluate \method{} across six VQA benchmarks covering diverse scenarios such as fine-grained perception, high-resolution grounding, and complex spatial and compositional reasoning.
Across these benchmarks, \method{} achieves the best overall performance among compared methods. 
In particular, \method{} consistently outperforms existing text-centric multi-agent VQA baselines. 
These results show that grounding multi-agent consensus in aligned visual evidence provides an effective mechanism for robust and reliable multimodal collaboration.


\paragraph{Contributions} Our contributions are threefold:
\begin{itemize}
\item \textbf{Key Insight:} We identify implicit visual evidence and evidence misalignment as key limitations of existing multi-agent VQA methods. Through a pilot study, we show that answer-level agreement alone is insufficient for reliable consensus; \textit{aligned visual evidence}---shared support from the image regions agents rely on---is essential, motivating our method.


    \item \textbf{Evidence-Centered Framework:} We propose \method{}, a training-free evidence-centered  multi-agent framework that makes agents' grounding 
    regions explicit as visual evidence. By leveraging evidence consistency, \method{} enables  diagnosis,  revision, and final arbitration based on what agents see rather than only on what they answer.

    

     \item \textbf{Extensive Evaluation:} We conduct comprehensive experiments across six VQA benchmarks spanning diverse visual reasoning tasks. Results show that \method{} achieves the best overall performance among compared methods and substantially outperforms text-centric multi-agent baselines.
\end{itemize}


\section{Related Work}

\paragraph{Agentic VQA.}

Agentic VQA methods aim to improve visual question answering by augmenting inference-time reasoning and coordination. 
Single-agent approaches include Chain-of-Thought~\citep{wei2022chain}, Zero-shot CoT~\citep{kojima2022large}, Self-Consistency~\citep{wang2022self}, iterative refinement~\citep{madaan2023self}, self-feedback~\citep{shinn2023reflexion}, and structured reasoning search~\citep{yao2023tree,besta2024graph}. 
Multi-agent systems further enhance reasoning by coordinating agents ~\citep{du2024improving,liang2024encouraging,li2023camel,qian2024chatdev,hong2024metagpt}, and recent methods incorporate perception agents, blind judges, or external tools to support answer revision for VQA~\citep{jiang2024multi,adhikari2025debating,sivakumaran2026dart}. 
Despite these advances, most approaches exchange only textual rationales without exposing the supporting visual regions, limiting evidence-grounded verification and revision. Tool-augmented methods such as DART~\cite{sivakumaran2026dart} offer stronger evidence access, but introduce additional tool dependency and deployment complexity.

\paragraph{Evidence-based Visual Reasoning.}

Evidence-based visual reasoning supports model predictions with identifiable evidence. 
Prior work mainly exploits three types of evidence: \textit{spatial evidence}, such as visual pointing and localized regions~\citep{park2018multimodal,wu2019faithful}, where grounding provides a natural way to make question-relevant regions explicit~\citep{chen2022grounding, reich2024role, reich2024uncovering, yi2025corgi, khayatkhoei2025mllms, man2025argus}; \textit{semantic evidence}, such as textual justifications, visual claims, or multimodal rationales~\citep{park2018multimodal, li2018vqa, wu2019faithful,li2025multimodal}; and \textit{process-level evidence}, where evidence is integrated through localize-before-answer, grounded chain-of-thought, or post-hoc verification~\citep{nguyen2025localizing,man2025argus,yi2025corgi}. 
Despite these advances, most methods use evidence only to support individual predictions, rather than to coordinate agents through shared visual evidence, leaving the complementary benefits of multi-agent reasoning underexplored.

\section{Pilot Study}
\label{sec:pilot}


We first conduct a pilot study on MMVP~\cite{tong2024eyes} to examine the role of aligned visual evidence in multi-agent VQA. 
Specifically, we use three heterogeneous VLM agents and ask each agent to independently predict an answer and output grounding boxes for its supporting evidence. We then group samples by two factors: whether agents reach unanimous answer agreement and whether their grounding boxes are spatially aligned (see Appendix~\ref{app:Pilot Study Setup} for more details).
Table~\ref{tab:pilot} reports the answer-accuracy  for each group.
Specifically, we investigate two questions:

\begin{table}[t]
\caption{\textbf{Pilot decomposition on MMVP.} Samples are partitioned by whether agents agree on the answer and whether their supporting evidence is spatially aligned. ``--'' denotes no samples in the corresponding group.}
\centering
\small
\setlength{\tabcolsep}{5pt}
\renewcommand{\arraystretch}{0.6}

\resizebox{\columnwidth}{!}{%
\begin{tabular}{lcc}
\toprule
\textbf{Answer}
& \multicolumn{2}{c}{\textbf{Evidence Consistency}} \\
\cmidrule(lr){2-3}
\textbf{Consistency}
& \textbf{Aligned}
& \textbf{Dispersed} \\
\midrule

\textbf{Consistent}
& \textbf{85.98\%} {\scriptsize (92/107)}
& 77.65\% {\scriptsize (66/85)} \\

\textbf{Inconsistent}
& -- {\scriptsize (0/0)}
& 52.78\% {\scriptsize (57/108)} \\
\bottomrule
\end{tabular}%
}

\label{tab:pilot}
\end{table}

\textbf{Q1: Is evidence alignment associated with answer consistency?}
Yes.
All evidence-aligned samples are also answer-consistent: the pilot contains 107 aligned-and-consistent cases and no aligned-but-inconsistent cases.
In contrast, evidence-dispersed samples include both answer-consistent and answer-inconsistent cases, with 85 and 108 samples respectively.
This suggests that {aligned grounding regions are strongly associated with answer agreement among agents}.

\textbf{Q2: Is answer agreement sufficient without aligned evidence?}
No.
Among answer-consistent samples, evidence-aligned cases achieve 85.98\% accuracy, while evidence-dispersed cases drop to 77.65\%.
When both answer and evidence are inconsistent, accuracy further decreases to 52.78\%.
Thus, answer consensus is substantially more reliable when supported by aligned visual evidence.

Overall, these results suggest that answer agreement alone is insufficient for trustworthy multi-agent VQA; agents must also verify that their predictions are supported by aligned visual evidence. 
This motivates our core principle: \textit{agents must see and align their visual evidence before agreeing.}


\begin{figure*}[t]
    \centering
    \includegraphics[width=\textwidth]{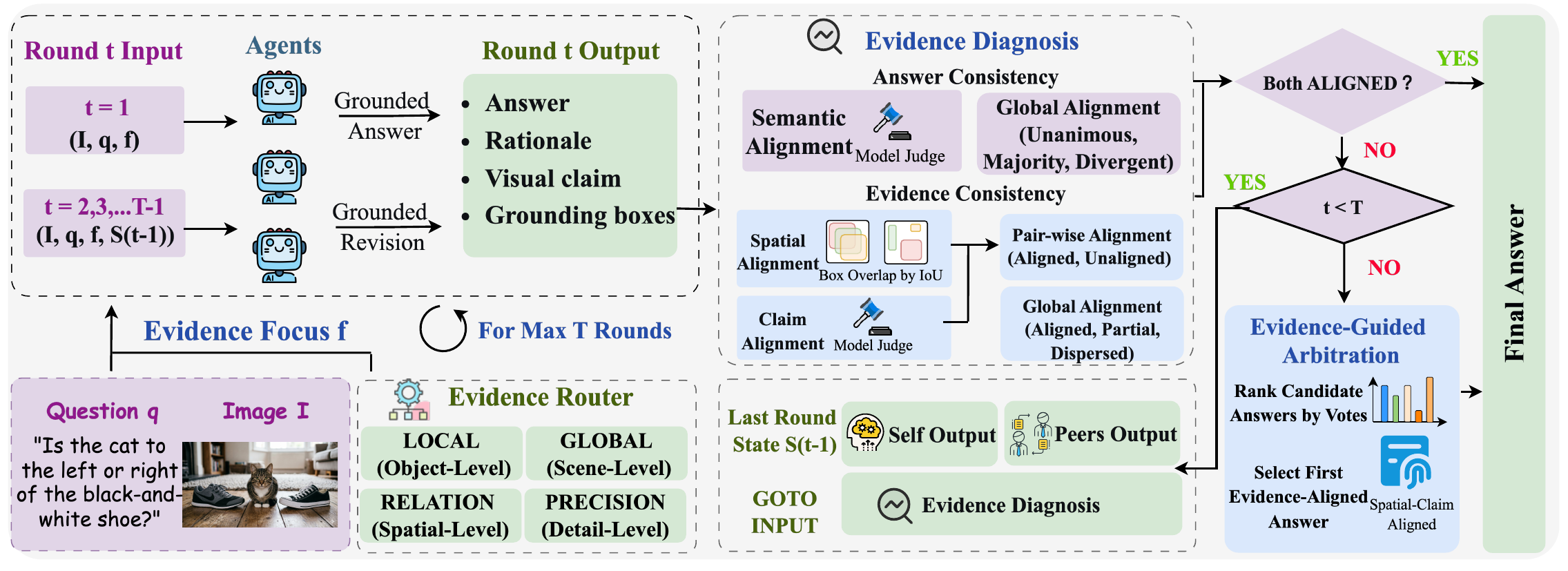}
    \caption{\textbf{Overview of \method{}}. The pipeline consists of five modules: 
    (1) \textbf{Evidence Routing}: guides grounding based on question type; 
    (2) \textbf{Grounded Answer}: agents generate initial answers with visual evidence, including grounding regions and visual claims explaining how the grounded regions support the answer; 
    (3) \textbf{Evidence Diagnosis}: evaluates consistency of answers and visual evidence across agents; 
    (4) \textbf{Grounded Revision}: agents revise predictions using previous-round state; 
    (5) \textbf{Evidence-Guided Arbitration}: ranks and selects final answers based on aligned evidence. 
   This design grounds multi-agent consensus in visual evidence, using cross-agent evidence consistency as a trustworthy signal to strengthen the final decision.}
    \label{fig:overview}
\end{figure*}



\section{Method}
\label{sec:method}

Guided by this insight, we propose \method{} 
(\textbf{E}vidence-\textbf{A}ligned \textbf{G}rounded mu\textbf{L}ti-agent r\textbf{E}asoning), 
a training-free evidence-centered multi-agent framework for VQA.
Instead of coordinating agents only through answers, \method{} makes each agent's supporting evidence explicit through grounding regions and visual claims, and uses evidence consistency as a control signal for accepting, revising, and arbitrating predictions.

As shown in Figure~\ref{fig:overview} and Algorithm~\ref{alg:eagle}, \method{} consists of five core modules. 
\textbf{Evidence Routing} determines the appropriate type of visual evidence and directs subsequent grounding. 
\textbf{Grounded Answer} asks agents to produce initial answers with explicit supporting evidence. 
\textbf{Evidence Diagnosis} evaluates answer agreement and evidence consistency. 
\textbf{Grounded Revision} guides agents to revise their predictions by re-examining the image conditioned on the previous-round state. 
\textbf{Evidence-Guided Arbitration} resolves remaining disagreements by prioritizing higher-voted answers supported by better-aligned evidence. 
More implementation details are provided in Appendix~\ref{app:method}.

\subsection{Evidence Routing}

Evidence Routing helps agents generate task-appropriate grounding evidence. 
Different VQA tasks require different grounding granularities: object-recognition questions may focus on a single local region, spatial-relation questions require related object pairs, and OCR questions require fine-grained text regions.
A generic instruction such as ``ground the evidence supporting your answer'' can lead to mismatched grounding boxes, including incomplete object sets, overly coarse regions, or boxes at inconsistent granularity. 
This makes evidence comparison unreliable.
Detailed failure examples are provided in Appendix~\ref{app:routing-examples}.

To address this, \method{} uses a  LLM-based Evidence Router. 
Given a question $q$, the router classifies it into a coarse evidence focus $f$: 
\textsc{Local} for object-level evidence, 
\textsc{Relation} for multi-object spatial evidence, 
\textsc{Global} for scene-level evidence, and 
\textsc{Precision} for fine-grained visual readouts. 
Based on the predicted focus, \method{} assigns a task-specific grounding instruction to guide both Grounded Answer and Grounded Revision. 
This makes the resulting visual evidence more effective and comparable across agents.



\subsection{Grounded Answer}

{Grounded Answer} is the first-round prediction, where each agent independently answers the question with explicit supporting visual evidence. 
Given the image $I$, question $q$, and routed evidence focus $f$, each agent $a_i \in \mathcal{A}$ produces:
\begin{equation}
\label{eq:grounded-answer-main}
o_i^{(1)} = a_i(I,q,f)
=
\left(y_i^{(1)}, r_i^{(1)}, b_i^{(1)}, c_i^{(1)}\right).
\end{equation}
Here, $y_i^{(1)}$ is the answer, $r_i^{(1)}$ is the rationale, $b_i^{(1)}$ denotes the grounding boxes, and $c_i^{(1)}$ is an atomic visual claim explaining how the grounded regions support the answer. 
Together, the grounding boxes and visual claims, capturing both spatial and semantic information, make the supporting evidence explicit for subsequent modules.

\subsection{Evidence Diagnosis}

{Evidence Diagnosis} is a post-processing stage that assesses the current output state through \emph{answer consistency} and \emph{evidence consistency}, and provides diagnostic signals for subsequent decision making.

\subsubsection{Answer Consistency}

Answer consistency measures whether agents produce semantically consistent answers. 
Since different agents may express the same answer in different surface forms, we use an LLM-as-judge to normalize answers by semantic equivalence. 
Let $\tilde{\mathcal{Y}}^{(t)}=\{\tilde{y}_i^{(t)}\}_{i=1}^{M}$ denote the set of semantic-normalized answers produced by $M$ agents at round $t$. 
We categorize answer consistency as:
\begin{equation}
C_y^{(t)} =
\begin{cases}
\textsc{Unanimous}, 
& |\mathrm{unique}(\tilde{\mathcal{Y}}^{(t)})|=1,\\
\textsc{Majority}, 
& \exists \tilde{y}:\mathrm{count}(\tilde{y}) > M/2,\\
\textsc{Divergent}, 
& \text{otherwise}.
\end{cases}
\end{equation}

\subsubsection{Evidence Consistency}

Evidence consistency measures whether agents rely on compatible visual evidence. 
We evaluate it using two complementary signals: \emph{spatial alignment} between grounding boxes and \emph{claim alignment} between visual claims. 
An agent pair is considered visually aligned only if both signals are satisfied, and a round is considered evidence-aligned only if all agent pairs are visually aligned.

\paragraph{Spatial alignment.}
Since the supporting evidence for a question may span multiple regions, we compare grounding-box sets using mutual coverage.
For an agent pair $(i,j)$, the directional coverage from $i$ to $j$ is defined as:
\begin{equation}
\mathrm{Cov}_{i\rightarrow j}^{(t)}
=
\frac{1}{|b_i^{(t)}|}
\sum_{b\in b_i^{(t)}}
\mathbb{I}
\left[
\max_{b'\in b_j^{(t)}} \mathrm{IoU}(b,b') > \tau_{\mathrm{iou}}
\right].
\end{equation}
where $\tau_{\mathrm{iou}}$ is a predefined IoU threshold.

The two agents are considered spatially aligned only if their box sets mutually cover each other:
\begin{equation}
\mathrm{Align}^{\mathrm{sp},(t)}_{ij}
=
\mathbb{I}
\left[
\mathrm{Cov}_{i\rightarrow j}^{(t)}=1
\land
\mathrm{Cov}_{j\rightarrow i}^{(t)}=1
\right].
\end{equation}

\paragraph{Claim alignment.}
Spatial overlap alone does not guarantee that agents interpret the evidence consistently. 
We therefore use an LLM judge to compare their visual claims conditioned on the question:
\begin{equation}
\mathrm{Align}^{\mathrm{cl},(t)}_{ij}
=
\mathrm{Judge}\big(q, c_i^{(t)}, c_j^{(t)}\big),
\end{equation}
where $\mathrm{Align}^{\mathrm{cl},(t)}_{ij}=1$ indicates that the two claims describe compatible visual evidence for answering question $q$, and $0$ indicates incompatible claims.

\paragraph{Evidence alignment.}
An agent pair is considered visually aligned only when both spatial and claim alignment hold:
\begin{equation}
\mathrm{Align}_{ij}^{(t)}
=
\mathbb{I}
\left[
\mathrm{Align}^{\mathrm{sp},(t)}_{ij}=1
\land
\mathrm{Align}^{\mathrm{cl},(t)}_{ij}=1
\right].
\end{equation}
Let $\mathcal{P}=\{(i,j)\mid 1\leq i<j\leq M\}$ denote the set of all agent pairs. 
We aggregate pairwise evidence alignment into round-level evidence consistency:
\begin{equation}
C_g^{(t)} =
\begin{cases}
\textsc{Aligned}, 
& \forall (i,j)\in\mathcal{P},~\mathrm{Align}_{ij}^{(t)}=1,\\
\textsc{Partial}, 
& \exists (i,j)\in\mathcal{P},~\mathrm{Align}_{ij}^{(t)}=1,\\
\textsc{Dispersed}, 
& \text{otherwise}.
\end{cases}
\end{equation}

\subsubsection{Evidence-Gated Early Exit}

Evidence-Gated Early Exit accepts the current round only when agents achieve both unanimous answer agreement and aligned visual evidence:
\begin{equation}
C_y^{(t)} = \textsc{Unanimous} \;\land\; C_g^{(t)} = \textsc{Aligned}.
\end{equation}
If this condition is not met and the maximum number of revision rounds has not been reached, \method{} proceeds to the next revision round, guided by the diagnostic summary
\begin{equation}
D^{(t)}
=
\left\{
C_y^{(t)},
C_g^{(t)},
\mathrm{Align}_{ij}^{(t)}
\right\},
\end{equation}
which provides an explicit signal of answer-level and evidence-level consistency. Otherwise, the process proceeds to final arbitration.



\subsection{Grounded Revision}

{Grounded Revision} refines predictions by encouraging agents to verify visual evidence before updating their predictions.
Each agent revises its prediction using the previous-round state as reference, including its own output, peer outputs, and diagnostic signals.
Specifically, each agent $a_i$ produces:
\begin{equation}
\label{eq:grounded-revision}
o_i^{(t)}
=
a_i\left(I, q, f, \mathcal{S}^{(t-1)}\right)
=
\left(
y_i^{(t)}, r_i^{(t)}, b_i^{(t)}, c_i^{(t)}
\right).
\end{equation}
Here, $\mathcal{S}^{(t-1)}=\{\mathcal{O}^{(t-1)}, \mathcal{D}^{(t-1)}\}$ denotes the previous-round state, containing the previous agent outputs and their diagnosis results.
\begin{table*}[t]
\caption{\textbf{Main results on six VQA benchmarks.}
We report average accuracy (Acc.). Best results are highlighted in bold, and second-best results are underlined. 
\method{} achieves the highest average accuracy and obtains the best or tied-best performance on five benchmarks, while remaining second-best on V*Bench.}
\centering
\scriptsize
\setlength{\tabcolsep}{3pt}
\renewcommand{\arraystretch}{0.2}

\resizebox{\textwidth}{!}{%
\begin{tabular}{llccccccc}
\toprule
\textbf{Method}
& \textbf{Agent}
& \textbf{GQA}
& \textbf{MME}
& \textbf{V*Bench}
& \textbf{WhatsUp}
& \textbf{MMVP}
& \textbf{MC-Bench}
& \textbf{Avg.} \\
\midrule

\multicolumn{9}{l}{\textbf{\textit{Single-agent methods}}} \\
\midrule

\multirow{3}{*}{Zero-shot CoT}
& \qwenlogo~Qwen
& 50.70 & 83.60 & 27.75 & 92.30 & 75.33 & 68.00 & 66.28 \\
& \glmlogo~GLM
& 52.00 & 78.50 & \textbf{33.51} & 90.30 & 76.00 & 53.00 & 63.89 \\
& \internvllogo~InternVL
& 49.20 & 78.00 & 28.80 & 63.00 & 69.00 & 62.40 & 58.40 \\

Self-Consistency
& \qwenlogo~Qwen
& 49.70 & 84.20 & 27.23 & 93.10 & 75.33 & 68.60 & 66.36 \\

Self-Refine
& \qwenlogo~Qwen
& 9.60 & 30.60 & 21.99 & 35.50 & 60.33 & 27.10 & 30.85 \\

\midrule
\multicolumn{9}{l}{\textbf{\textit{Multi-agent methods}}} \\
\midrule

Debate(Vote)
& \qwenlogo~\glmlogo~\internvllogo
& 57.50 & \underline{85.20} & 28.27 & \underline{96.40} & 77.67 & 72.80 & 69.64 \\

Debate(Judge)
& \qwenlogo~\glmlogo~\internvllogo
& 57.20 & 84.80 & 25.65 & 96.20 & 78.00 & 72.60 & 69.08 \\

ReConcile
& \qwenlogo~\glmlogo~\internvllogo
& 56.80 & 84.90 & 27.23 & 95.90 & 77.00 & \underline{73.60} & 69.24 \\

DART
& \qwenlogo~\glmlogo~\internvllogo
& \underline{57.70} & 84.40 & 29.84 & 95.50 & \textbf{80.33} & 73.10 & \underline{70.15} \\

\textbf{\method{}}
& \qwenlogo~\glmlogo~\internvllogo
& \textbf{60.30} & \textbf{87.90} & \underline{32.98} & \textbf{99.20} & \textbf{80.33} & \textbf{75.40} & \textbf{72.69} \\
\bottomrule
\end{tabular}%
}

\label{tab:main-results}
\end{table*}

\begin{table}[t]
\caption{\textbf{Efficiency comparison of multi-agent methods.}
Acc. denotes average accuracy, Calls denotes the average number of VLM-agent calls per sample, Ratio denotes Acc./Calls, and Tools indicates whether external visual tools are used. 
\method{} achieves the highest efficiency without external visual tools.}
\centering
\scriptsize
\setlength{\tabcolsep}{3.5pt}
\renewcommand{\arraystretch}{0.6}

\resizebox{\columnwidth}{!}{%
\begin{tabular}{lcccc}
\toprule
\textbf{Method}
& \textbf{Acc. $\uparrow$}
& \textbf{Calls $\downarrow$}
& \textbf{Ratio $\uparrow$}
& \textbf{Tools} \\
\midrule

Debate (Vote)  
& 69.64 & 4.286 & 16.25 & \xmark \\

Debate (Judge) 
& 69.08 & 5.278 & 13.09 & \xmark \\

ReConcile      
& 69.24 & 4.343 & 15.94 & \xmark \\

DART           
& 70.15 & 5.322 & 13.18 & \cmark \\

\textbf{\method{}} 
& \textbf{72.69} & \textbf{4.087} & \textbf{17.79} & \xmark \\

\bottomrule
\end{tabular}%
}

\label{tab:efficiency}
\end{table}

\subsection{Evidence-Guided Arbitration}

{Evidence-Guided Arbitration} resolves final disagreements by favoring answers with higher votes and aligned evidence.

Let $\mathcal{G}(y)=\{i \mid y_i^{(T)}=y\}$ denote the agents predicting answer $y$ in the final round. 
Candidates are checked in descending order of $|\mathcal{G}(y)|$, and \method{} selects the first evidence-aligned answer:
\begin{equation}
\hat{y}
=
\operatorname*{first}_{y}
\left[
|\mathcal{G}(y)|>1
\land
C_g(\mathcal{G}(y))=\textsc{Aligned}
\right].
\end{equation}
If no evidence-aligned answer group exists, EAGLE falls back to the highest-voted answer; ties are broken by the strongest grounded single-agent prediction.
This reflects the core intuition that explicit visual support is preferable to unsupported or evidence-inconsistent agreement.

\section{Experiment}
\label{sec:exp}
\subsection{Experimental Setup}
\label{sec:setup}

\paragraph{Models.}
Following ReConcile~\cite{chen2024reconcile}, we use $M=3$ agents and instantiate them with a heterogeneous committee of open-source VLMs: Qwen3-VL-32B~\cite{bai2025qwen3}, GLM-4.6V-Flash~\cite{hong2025glm}, and InternVL3.5-38B~\cite{wang2025internvl3}, which encourages diverse and complementary visual interpretations.

\paragraph{Baselines.}
We compare \method{} with representative agentic baselines, including single-agent methods Zero-shot CoT~\citep{kojima2022large}, Self-Consistency~\citep{wang2022self}, and Self-Refine~\citep{madaan2023self}, as well as multi-agent methods Multi-Agent Debate~\citep{du2024improving,liang2024encouraging}, ReConcile~\citep{chen2024reconcile}, and DART~\citep{sivakumaran2026dart}. 
For Zero-shot CoT, we report each individual agent to reflect its inherent capability; for other single-agent baselines, we use Qwen3-VL as the backbone, since it is the strongest individual model in our setting. 
All multi-agent baselines use the same three-agent committee for fair comparison. 
For Multi-Agent Debate, we evaluate two final aggregation variants: vote-based aggregation and judge-based aggregation. 
More details are provided in Appendix~\ref{app:baselines}.

\paragraph{Benchmarks.}

We evaluate \method{} on six diverse VQA benchmarks where relevant visual cues are essential for correct answering: V*Bench~\cite{wu2024v}, WhatsUp~\cite{kamath2023s}, MME~\cite{fu2026mme}, GQA~\cite{hudson2019gqa}, MC-Bench~\cite{xu2025mc}, and MMVP~\cite{tong2024eyes}. 
Together, these benchmarks assess the generality of \method{} across diverse visual understanding scenarios. 

\paragraph{Evaluation.}


We report accuracy as the main metric. 
Following the LLM-as-a-judge protocol~\citep{manas2024improving}, we use the lightweight LLaMA-3.2-1B-Instruct~\citep{grattafiori2024llama} to measure final-answer correctness. 
Within the pipeline, the same lightweight model is used for evidence routing and answer/claim consistency checking. 
To verify reliability, we manually inspect 100 randomly sampled examples from the final-answer evaluation and each judge-based pipeline component, observing strong agreement with human annotations (average Cohen's $\kappa=0.94$).


\subsection{Main Results}
\label{sec:main-results}

\paragraph{Overall performance.}
Table~\ref{tab:main-results} reports the main results on six VQA benchmarks.
\method{} achieves the best overall performance among all baselines, obtaining the best or tied-best results on five benchmarks and the second-best result on V*Bench.
Compared with the strongest baseline, DART, which uses external visual tools to resolve agent disagreements, \method{} improves average accuracy by 2.54 percentage points.
These results demonstrate the effectiveness of using explicit visual evidence to coordinate multi-agent consensus.

\paragraph{Comparison with single-agent inference.}
\method{} substantially outperforms single-agent inference. 
It improves over the strongest single-agent baseline, Self-Consistency with Qwen3-VL, by 6.33 percentage points on average, and surpasses the best zero-shot individual agent by 6.41 points. 
The consistent gains across benchmarks show that evidence-aware multi-agent coordination provides benefits beyond the capability of any single model.

\paragraph{Comparison with multi-agent baselines.}
\method{} achieves the highest average performance among methods under the same three-agent committee, improving over Debate (Vote), Debate (Judge), ReConcile, and tool-augmented DART by 3.05, 3.61, 3.45, and 2.54 points, respectively.
Unlike these baselines, which rely on answers, rationales, confidence, or external tools, \method{} directly compares the explicit visual evidence supporting each answer, demonstrating that evidence-level coordination provides a promising mechanism to strengthen multi-agent collaboration.

\subsection{Efficiency Analysis}
\label{sec:efficiency}

\paragraph{Evidence-guided interaction improves efficiency.}
Table~\ref{tab:efficiency} compares multi-agent methods in terms of average accuracy, VLM-agent calls, and ratio (Acc./Calls). 
\method{} achieves the highest average accuracy while using the fewest VLM-agent calls, yielding the best ratio of 17.79 among all methods. 
Compared with DART, \method{} reduces calls by 23.2\% while improving accuracy from 70.15 to 72.69, all without relying on external tools.

\paragraph{Gains do not come from longer discussion.}
The efficiency of EAGLE arises from its evidence-aware revision rather than from longer discussions.
The parameter ablations in Figure~\ref{fig:parameter-ablation} show that one grounded revision round (R2) is sufficient for reliable consensus, whereas additional rounds may cause over-revision and reduce performance.

\subsection{Ablation Study}
\label{sec:ablation-analysis}

\begin{table}[t]
\caption{\textbf{Pipeline component ablations.}
Each row removes one component from the full pipeline, and downward arrows indicate the absolute performance drops from the full setting.}
\centering
\small
\setlength{\tabcolsep}{5pt}
\renewcommand{\arraystretch}{1.12}

\resizebox{\columnwidth}{!}{%
\begin{tabular}{lccccc}
\toprule
\textbf{Configuration}
& \textbf{GQA}
& \textbf{MME}
& \textbf{WhatsUp}
& \textbf{MC-Bench}
& \textbf{Avg.} \\
\midrule

\textbf{Full}
& \textbf{60.3} & \textbf{87.9} & \textbf{99.2} & \textbf{75.4} & \textbf{80.70} \\

\midrule
w/o Routing
& 58.2\dn{2.1} & 84.6\dn{3.3} & 94.3\dn{4.9} & 73.0\dn{2.4} & 77.52\dn{3.18} \\

w/o Grounding
& 57.3\dn{3.0} & 84.7\dn{3.2} & 97.2\dn{2.0} & 70.6\dn{4.8} & 77.45\dn{3.25} \\

w/o Arbitration
& 58.8\dn{1.5} & 86.5\dn{1.4} & 97.9\dn{1.3} & 74.0\dn{1.4} & 79.30\dn{1.40} \\

w/o Early Exit
& 58.9\dn{1.4} & 86.5\dn{1.4} & 98.1\dn{1.1} & 73.3\dn{2.1} & 79.20\dn{1.50} \\

\bottomrule
\end{tabular}%
}

\label{tab:ablation-pipeline}
\end{table}

\begin{table}[t]
\caption{\textbf{Revision-context ablations.}
Each row removes one component from the full pipeline, and downward arrows indicate the absolute performance drops from the full setting.}
\centering
\small
\setlength{\tabcolsep}{5pt}
\renewcommand{\arraystretch}{1.12}

\resizebox{\columnwidth}{!}{%
\begin{tabular}{lccccc}
\toprule
\textbf{Configuration}
& \textbf{GQA}
& \textbf{MME}
& \textbf{WhatsUp}
& \textbf{MC-Bench}
& \textbf{Avg.} \\
\midrule

\textbf{Full}
& \textbf{60.3} & \textbf{87.9} & \textbf{99.2} & \textbf{75.4} & \textbf{80.70} \\

\midrule
w/o Self Out
& 57.0\dn{3.3} & 85.3\dn{2.6} & 97.6\dn{1.6} & 73.4\dn{2.0} & 78.32\dn{2.38} \\

w/o Peer Out
& 58.4\dn{1.9} & 86.1\dn{1.8} & 97.5\dn{1.7} & 72.8\dn{2.6} & 78.70\dn{2.00} \\

w/o Diagnosis
& 58.1\dn{2.2} & 85.4\dn{2.5} & 98.7\dn{0.5} & 73.2\dn{2.2} & 78.85\dn{1.85} \\

\bottomrule
\end{tabular}%
}

\label{tab:ablation-r2-context}
\end{table}

\begin{figure}[t]
    \centering
    \includegraphics[width=\columnwidth]{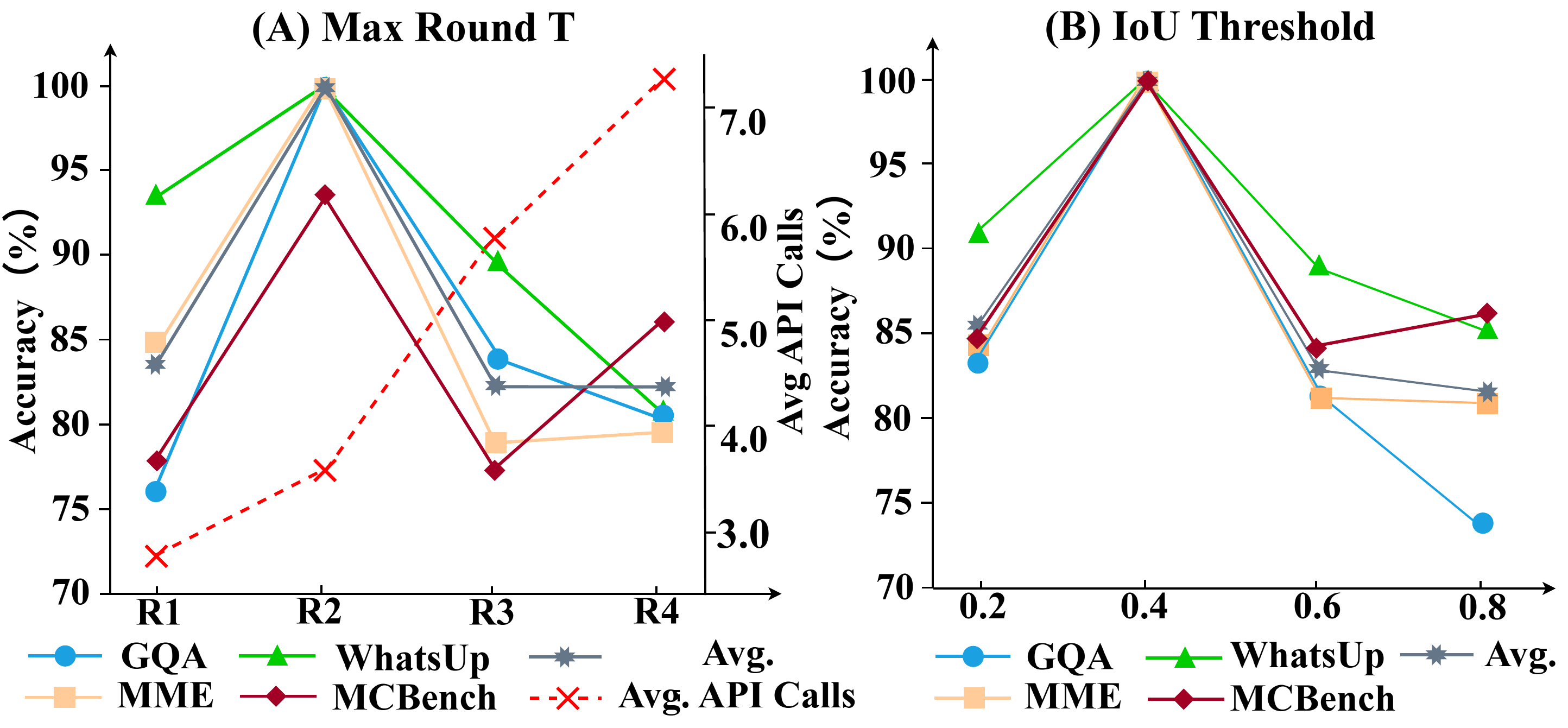}
    \vspace{-2mm}
    \caption{\textbf{Parameter ablations.}
    (A) Effect of the maximum number of revision rounds $T$, showing that one grounded revision is sufficient for reliable consensus;
    (B) sensitivity to IoU threshold $\tau_{\mathrm{iou}}$, with 0.4 providing the best spatial alignment across agents.}
    \label{fig:parameter-ablation}
\end{figure}

\definecolor{casegray}{HTML}{F5F6F8}
\definecolor{caseblue}{HTML}{EAF2FF}
\definecolor{casegreen}{HTML}{EAF7EA}
\definecolor{casered}{HTML}{FDECEC}

\newcommand{\glmicon}{\includegraphics[height=1.15em]{fig/glmv-color.pdf}}
\newcommand{\qwenicon}{\includegraphics[height=1.15em]{fig/qwen-color.pdf}}
\newcommand{\internvlicon}{\includegraphics[height=1.15em]{fig/internlm-color.pdf}}

\begin{table*}[t]
\caption{
\textbf{A V*Bench case of \method{}.}
R1 and R2 denote the initial grounded answer and grounded revision. 
R1 grounding boxes are color-coded by agent: GLM in green, Qwen in red, and InternVL in blue.
Guided by peers' visual evidence, the initially incorrect Qwen corrects its answer, leading to consensus on the correct backpack color.
}
\centering
\scriptsize
\setlength{\tabcolsep}{2pt} 
\renewcommand{\arraystretch}{0.9} 

\begin{tabular}{p{0.35\textwidth}p{0.62\textwidth}} 

\begin{minipage}[t]{\linewidth}
\vspace{0pt}
\includegraphics[width=\linewidth,height=2.5cm,keepaspectratio]{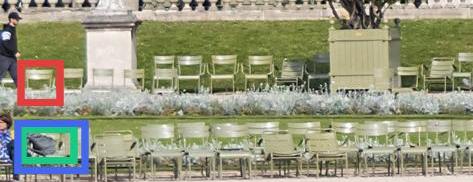}

\vspace{0.5em}
\raggedright
\textbf{Question:} What is the color of the backpack carried by the person in the left corner?

\textbf{Options:} A = gray (GT), B = white, C = yellow, D = red.
\end{minipage}
&
\begin{minipage}[t]{\linewidth}
\vspace{0pt}
\centering
\scriptsize 
\setlength{\tabcolsep}{1pt}
\renewcommand{\arraystretch}{0.88}

\begin{tabular}{p{0.12\linewidth}p{0.42\linewidth}p{0.42\linewidth}} 
\toprule
\multicolumn{1}{c}{\textbf{Agent}} & 
\multicolumn{1}{c}{\textbf{R1: Grounded Answer}} & 
\multicolumn{1}{c}{\textbf{R2: Grounded Revision}} \\
\midrule

\rowcolor{white} 
\centering
\raisebox{-1\height}{\glmicon} 
&
\textbf{Answer:} A (gray) \newline
\textit{Reasoning:} Grounds answer in clearly visible backpack.
&
\textbf{Answer:} A (gray) \newline
\textit{Reasoning:} Keeps original grounding; confirms correct color.
\\

\rowcolor{white}
\centering
\raisebox{-1\height}{\qwenicon} 
&
\textbf{Answer:} B (white) \newline
\textit{Reasoning:} Mistakenly identifies the white chair above the person as the backpack.
&
\textbf{Answer:} A (gray) \newline
\textit{Reasoning:} Revises to gray by referencing peers’ grounding.
\\

\rowcolor{white}
\centering
\raisebox{-1\height}{\internvlicon}
&
\textbf{Answer:} A (gray) \newline
\textit{Reasoning:} Grounds answer in clearly visible backpack.
&
\textbf{Answer:} A (gray) \newline
\textit{Reasoning:} Maintains correct grounding and color.
\\

\bottomrule
\end{tabular}
\end{minipage}
\\

\end{tabular}

\label{tab:main-case-vstar75}
\end{table*}


\paragraph{Pipeline Component ablations.}
Table~\ref{tab:ablation-pipeline} reports ablations of the major components in \method{}. 
\textit{w/o Routing} removes Evidence Routing; \textit{w/o Grounding} removes explicit grounding boxes and keeps only textual visual descriptions; \textit{w/o Arbitration} replaces evidence-guided arbitration with vote-based selection; and \textit{w/o Early Exit} forces all samples to pass through the full pipeline.
Performance drops after removing any component, showing that each module contributes to the full pipeline. 
The largest drops come from removing Evidence Routing or explicit grounding boxes, suggesting that task-appropriate grounding and explicit visual evidence are crucial for reliable prediction. 
Removing evidence-guided arbitration also hurts performance, since unresolved disagreements are no longer verified by aligned visual evidence. 
Disabling Early Exit leads to additional degradation, indicating that unnecessary revision after evidence-aligned consensus can introduce over-revision.

\paragraph{Revision context ablations.}
Table~\ref{tab:ablation-r2-context} examines which information is essential during Grounded Revision. 
\textit{w/o Self Out} removes the agent's own previous output, \textit{w/o Peer Out} removes other agents' outputs, and \textit{w/o Diagnosis} removes evidence diagnosis signals.  
Removing any of these inputs consistently reduces performance, indicating that Grounded Revision is not a second attempt at answering the question. 
Instead, it is a structured evidence-guided re-examination process: self outputs preserve each agent's prior visual evidence, peer outputs provide alternative grounded references, and diagnostic signals reveal whether disagreement arises from answers, evidence, or both.

\paragraph{Parameter ablations.}
\label{sec:parameter-ablations}

Figure~\ref{fig:parameter-ablation} evaluates the two main experimental parameters: the maximum number of revision rounds and the IoU threshold for spatial alignment. 
R2 achieves the best macro-average accuracy, indicating that a single grounded revision round is sufficient, while additional rounds may cause over-revision and unnecessary computation. 
For spatial alignment, $\tau_{\mathrm{iou}}=0.4$ performs best: a lower threshold may align weakly related regions, whereas a higher threshold may discard valid evidence due to small box shifts across agents.

\subsection{Case Study}
\label{sec:case-study-main}

Table~\ref{tab:main-case-vstar75} illustrates how \method{} corrects an initially unsupported prediction through grounded revision. 
The question asks for the backpack color of the person in the left corner. 
GLM and InternVL correctly answer A (gray) by grounding on the backpack, whereas Qwen answers B (white) after mistakenly including a nearby white chair. 
Through evidence-aware interaction, Qwen re-examines the image, discards the irrelevant chair, and revises its answer to A, leading all agents to an evidence-aligned consensus. 
This case shows that \method{} improves multi-agent reasoning by aligning visual evidence rather than simply aggregating answer votes. 
Additional examples are provided in Appendix~\ref{app:case-study}.

\section{Conclusion}

We present \method{}, a training-free, evidence-centered multi-agent framework for VQA. 
\method{} makes agents' grounding regions and visual claims explicit, using evidence consistency to guide diagnosis, revision, and arbitration beyond answer-level agreement. 
Experiments on six VQA benchmarks show that \method{} achieves the best overall performance among compared methods while remaining lightweight and interpretable. 
These results highlight aligned visual evidence as an effective foundation for reliable multi-agent visual reasoning.

\section*{Limitations}

This work focuses on evidence-aligned multi-agent reasoning for static-image VQA, and several promising directions remain for future exploration. 
First, our current study mainly evaluates \method{} on English VQA benchmarks. 
Extending evidence-aligned coordination to multilingual VQA would further test its robustness across different linguistic and cultural contexts. 
Second, this work primarily considers image-based visual reasoning. 
Applying the same principle to broader multimodal scenarios, such as video understanding, embodied perception, document intelligence, and open-ended multimodal generation, is an important future direction. 
Third, while \method{} uses explicit grounding regions and visual claims to improve interpretability, future work may explore richer forms of evidence representation, such as temporal evidence, hierarchical scene structures, or cross-modal evidence chains. 
We hope this work can serve as a step toward more general evidence-grounded collaboration among multimodal agents.

\section*{Ethical Considerations}

This work studies a training-free multi-agent framework for visual question answering using existing VLMs and public VQA benchmarks. 
All public models and benchmarks are used only for research evaluation, and we do not redistribute any data or model weights. 
We do not introduce new human-subject data, private user data, or personally identifiable information. 
The proposed method aims to improve VQA reliability and interpretability by making supporting visual evidence explicit and comparable across agents. 
However, it should not replace human judgment in high-stakes scenarios such as medical, legal, or safety-critical decision-making. 
Although evidence alignment improves transparency, errors may still arise from incorrect visual perception, routing, or consistency checking. 
Sensitive applications should therefore include human oversight, careful validation, and domain-specific risk assessment.


\bibliography{latex/main}
\clearpage
\newpage
\appendix

\section*{Appendix}
\startcontents[sections]
\printcontents[sections]{l}{1}{\setcounter{tocdepth}{3}}

\section{Pilot Study Setup}
\label{app:Pilot Study Setup}

We conduct the pilot study on MMVP~\citep{tong2024eyes}, a benchmark emphasizing fine-grained visual evidence, using three heterogeneous VLM agents: Qwen3-VL, GLM-4.6V, and InternVL3.5. 
Each agent independently outputs an answer and grounding boxes for its supporting evidence. 
We define \emph{answer consistency} as unanimous agreement among the three agents, and use majority vote as the final prediction when answers are inconsistent. 
We define \emph{evidence alignment} as the case where all agent pairs have grounding boxes overlapping above an IoU threshold of $\tau=0.4$. 

Based on answer consistency and evidence alignment, we partition samples into four groups and report final-answer accuracy using ground-truth annotations. 
Table~\ref{tab:pilot} summarizes the results.

\section{Method}
\label{app:method}

\subsection{Overall Algorithm}
\label{app:Overall Algorithm}
To provide a comprehensive and structured view of the execution pipeline, we present the formal pseudocode of the proposed \method{} framework in Algorithm~\ref{alg:eagle}. Specifically, the algorithm formalizes the coordination among multiple visual-language agents ($\mathcal{A}$) by sequentially orchestrating the five modules detailed in Figure~\ref{fig:overview}: 
(1) \textsc{EvidenceRoute} guides grounding based on question type; 
(2) \textsc{Answer} prompts agents to generate initial answers with visual evidence, including grounding regions and visual claims; 
(3) \textsc{Diagnose} evaluates the consistency of answers and visual evidence across agents; 
(4) \textsc{Revise} allows agents to self-correct and revise predictions using the previous-round state; and 
(5) \textsc{Arbitrate} ranks and selects final answers based on aligned evidence when a deadlock occurs. 
By translating these modules into an algorithmic workflow, \method{} explicitly grounds multi-agent consensus in visual evidence, leveraging cross-agent consistency as a trustworthy signal.

\begin{tcolorbox}[
    float=t,
    title={Algorithm~\refstepcounter{algorithm}\label{alg:eagle}\thealgorithm\quad \method{} Framework},
    colback=lighterblue,
    colframe=tableblue,
    coltitle=white,
    colbacktitle=tableblue,
    fonttitle=\bfseries,
    boxrule=0.7pt,
    arc=2pt,
    left=4pt,
    right=4pt,
    top=4pt,
    bottom=4pt
]
\small
\begin{algorithmic}[1]
\Require Image $I$, question $q$, agents $\mathcal{A}$, max rounds $T$
\Ensure Final answer $\hat{y}$

\State $f \gets \textsc{EvidenceRoute}(q)$

\For{$t = 1$ \textbf{to} $T$}
    \If{$t = 1$}
        \State $\mathcal{O}^{(t)} \gets \textsc{Answer}(I,q,f,\mathcal{A})$
    \Else
        \State $\mathcal{O}^{(t)} \gets \textsc{Revise}(I,q,f,\mathcal{A},\mathcal{S}^{(t-1)})$
    \EndIf

    \State $\mathcal{D}^{(t)} \gets \textsc{Diagnose}(\mathcal{O}^{(t)})$

    \If{$\textsc{BothAligned}(\mathcal{D}^{(t)})$}
        \State $\hat{y} \gets \textsc{CommonAnswer}(\mathcal{O}^{(t)})$
        \State \textbf{break}
    \EndIf

    \State $\mathcal{S}^{(t)} \gets \{\mathcal{O}^{(t)}, \mathcal{D}^{(t)}\}$
\EndFor

\If{$\hat{y}$ is undefined}
    \State $\hat{y} \gets \textsc{Arbitrate}(\mathcal{S}^{(T)})$
\EndIf

\State \Return $\hat{y}$

\end{algorithmic}
\end{tcolorbox}

\subsection{Evidence Routing}
\label{app:routing-examples}

\subsubsection{Necessity: Why Routing is Needed}

Generic grounding instructions can fail because different VQA questions require different evidence formats. 
When all questions share the same broad instruction, agents may produce grounding regions with incompatible granularity or focus on different visual entities. 
This makes Evidence Diagnosis unreliable, since the system may compare boxes or claims that do not correspond to the same visual basis. 
We summarize representative failure patterns below.

\begin{itemize}
    \item \textsc{Local}. 
    Object- or attribute-level questions require evidence centered on the queried object and its relevant attribute. 
    For example, for ``What color is the car?'', a generic instruction may cause one agent to ground the whole street scene, another to ground several nearby vehicles, and another to ground only a small part of the car. 
    These regions are all visually plausible, but they do not provide equally useful evidence for the target attribute.

    \item \textsc{Relation}. 
    Spatial or interaction questions require evidence covering all entities involved in the relation. 
    For example, for ``Where is the cat relative to the shoes?'', one agent may ground only the cat, another may ground only the shoes, and another may ground a visually similar but incorrect shoe-like region. 
    Such outputs are difficult to compare because the relation is supported only when both the target and reference objects are localized.

    \item \textsc{Global}. 
    Scene-level questions require evidence from the overall layout rather than a single local object. 
    For example, for ``Is this scene indoors or outdoors?'', grounding a chair, a window, or a person may each be locally valid but insufficient for the global scene judgment. 
    The answer depends on broader cues such as layout, background, lighting, and environment context.

    \item \textsc{Precision}. 
    OCR, counting, and fine-grained readout questions require precise evidence. 
    For OCR questions such as ``What text is written on the road?'', grounding the whole road or crosswalk region is too coarse; the evidence should localize the specific painted word or text line. 
    For counting questions such as ``How many pedestrians are within 45 meters?'', agents should ground the counted instances individually rather than a broad street or crowd region. 
    Otherwise, one agent may count ambiguous distant figures while another counts only clearly visible instances, leading to inconsistent evidence and unstable revision.
\end{itemize}

These examples show that routing is not intended to solve the task directly. 
Instead, it provides an evidence-format prior that encourages agents to return comparable visual claims and grounding regions, making subsequent diagnosis, revision, and arbitration more stable.

\subsubsection{Implementation: LLM-Based Evidence Router}

Evidence Routing is implemented with an LLM-based router that predicts the evidence focus required by each question. 
Given only the question $q$, the router classifies it into one of four coarse evidence focuses:
\[
\textsc{Local},\quad
\textsc{Relation},\quad
\textsc{Global},\quad
\textsc{Precision}.
\]
The router does not take the image, candidate answers, agent responses, or grounding boxes as input. 
Therefore, it is not used to solve the VQA task directly. 
Instead, it only provides a question-level prior about what type of visual evidence should be grounded.

Specifically, \textsc{Local} is used for object-level evidence, such as questions about the category, color, shape, state, existence, or local attribute of a specific object. 
\textsc{Relation} is used for multi-object spatial, interaction, or comparative evidence, where agents need to inspect more than one entity and their configuration. 
\textsc{Global} is used for scene-level evidence, such as questions about the overall scene, layout, main activity, weather, mood, or distributed visual cues. 
\textsc{Precision} is used for fine-grained visual readouts, including OCR, counting, numbers, labels, symbols, measurements, chart/table reading, and other exact visual references.

After obtaining the predicted evidence focus $f$, \method{} maps it to a short task-specific grounding instruction. 
This instruction is appended to both the Grounded Answer prompt and the Grounded Revision prompt. 
The instruction does not replace the original VQA prompt; it only guides agents to produce visual claims and grounding boxes with a more appropriate granularity. 
This makes the resulting evidence more effective and more comparable across agents.

\paragraph{Routing Judge Prompt.}
We use the following prompt for LLM-based evidence routing.

\begin{tcolorbox}[
    colback=blue!6,
    colframe=blue!65!black,
    colbacktitle=blue!70!black,
    coltitle=white,
    title=\textbf{Evidence Routing Prompt},
    fonttitle=\bfseries,
    boxrule=0.6pt,
    arc=3pt,
    left=8pt,
    right=8pt,
    top=7pt,
    bottom=7pt,
    breakable
]
\small
\ttfamily
You are an evidence router for a visual question answering system.

\vspace{0.5em}

Your task is to decide what type of visual evidence an agent should focus on when answering the question.

\vspace{0.5em}

Given a question, classify it into exactly one of the following evidence focuses:

\vspace{0.5em}

\textbf{Local}:  
Choose this label when the question can usually be answered by inspecting one decisive object, attribute, or local visual detail.  
Examples include object category, color, shape, state, existence, or a local attribute of a specific object.

\vspace{0.5em}

\textbf{Relation}:  
Choose this label when the question requires evidence about multiple entities and their relationship.  
Examples include spatial relations, relative positions, interactions, comparisons, or configurations between objects.

\vspace{0.5em}

\textbf{Global}:  
Choose this label when the question requires understanding the whole image or distributed scene-level cues.  
Examples include the overall scene, global layout, main activity, scene category, weather, mood, common property, or image-level description.

\vspace{0.5em}

\textbf{Precision}:  
Choose this label when the question requires exact visual readout or fine-grained lookup.  
Examples include OCR, counting, numbers, labels, symbols, measurements, distances, angles, chart/table reading, or precise text/mark recognition.

\vspace{0.5em}

Important rules:
\begin{itemize}
    \item Only decide the evidence focus required for reliable grounding.
    \item Output exactly one label from: Local, Relation, Global, Precision.
\end{itemize}

\vspace{0.5em}

Question: \{question\}

\vspace{0.5em}

Output format:

\{
"evidence\_focus": "Local | Relation | Global | Precision"
\}
\end{tcolorbox}

\paragraph{Prompt Instantiation.}
Based on the predicted evidence focus, we append the corresponding grounding instruction to the agent prompt. 
The instruction specifies not only what visual evidence the agent should inspect, but also what regions should be grounded with bounding boxes. 
This helps agents produce task-appropriate visual claims and comparable grounding boxes.

\begin{tcolorbox}[
    colback=blue!6,
    colframe=blue!65!black,
    colbacktitle=blue!70!black,
    coltitle=white,
    title=\textbf{Additional Grounding Instruction},
    fonttitle=\bfseries,
    boxrule=0.6pt,
    arc=3pt,
    left=8pt,
    right=8pt,
    top=7pt,
    bottom=7pt,
    breakable
]
\small
\ttfamily
\textbf{Grounding instruction}:

\vspace{0.5em}

\textbf{local}: 
Focus on the single decisive object, attribute, or local detail. 
Ground the target object or the specific object part that directly supports the answer. 
Do not ground the whole scene or irrelevant surrounding regions.

\vspace{0.5em}

\textbf{relation}: 
Focus on all entities involved in the queried relation. 
Ground each necessary entity, including both the target object and the reference object, so that their spatial, interactive, or comparative relation can be verified. 
Do not ground only one side of the relation.

\vspace{0.5em}

\textbf{global}: 
Focus on the overall scene or distributed visual cues. 
If the answer depends on the whole image, ground the full scene or the broad scene region. 
If the answer depends on multiple distributed cues, ground the relevant regions jointly rather than selecting a single local object.

\vspace{0.5em}

\textbf{precision}: 
Focus on exact visual readouts such as text, numbers, labels, symbols, measurements, or counts. 
Ground the precise visual elements used for the answer: for OCR, box the exact word or text line; for counting, box the counted instances; for charts or diagrams, box the relevant labels, ticks, marks, or plotted elements. 
Avoid overly coarse boxes.
\end{tcolorbox}

\subsection{Grounded Answer}
\label{app:grounded-answering}

Grounded Answer is the first-round prediction, where each agent independently answers the question with explicit supporting visual evidence before observing other agents' outputs. 
Given the image $I$, question $q$, and routed evidence focus $f$, each agent $a_i \in \mathcal{A}$ produces:
\begin{equation}
\label{eq:grounded-answer}
o_i^{(1)}
=
a_i(I,q,f)
=
\left(
y_i^{(1)}, r_i^{(1)}, b_i^{(1)}, c_i^{(1)}
\right).
\end{equation}

where $y_i^{(1)}$ is the answer, $r_i^{(1)}$ is the rationale, $b_i^{(1)}$ denotes the grounding boxes, and $c_i^{(1)}$ is an atomic visual claim explaining how the grounded regions support the answer. 
Together, the grounding boxes and visual claims capture both spatial and semantic information, making the supporting evidence explicit for subsequent modules.

For each sample, the grounded-answering prompt contains the question $q$, the routed evidence focus $f$, and a focus-specific evidence-grounding instruction.

\begin{tcolorbox}[
    colback=blue!6,
    colframe=blue!65!black,
    colbacktitle=blue!70!black,
    coltitle=white,
    title=\textbf{Grounded Answer Prompt},
    fonttitle=\bfseries,
    boxrule=0.6pt,
    arc=3pt,
    left=8pt,
    right=8pt,
    top=7pt,
    bottom=7pt,
    breakable
]
\begin{lstlisting}[
    basicstyle=\ttfamily\small,
    columns=fullflexible,
    breaklines=true,
    breakatwhitespace=true,
    keepspaces=true,
    showstringspaces=false,
    frame=none
]
You are an independent visual question-answering agent.
The image is the only source of visual evidence.

QUESTION:
{question}

Evidence focus: {mode}
Additional evidence-grounding instruction:
{additional_evidence_grounding_instruction}

Task:
1. Inspect the image independently.
2. Follow the evidence-grounding instruction.
3. Answer the question with concise evidence-grounded reasoning.
4. Provide one atomic visual claim that directly supports your answer.
5. Provide the grounding boxes for the image region(s) that support this visual claim.

Output JSON schema:
{
  "reasoning": "brief evidence-grounded reasoning",
  "visual_claim": "one atomic visual finding explaining how the grounded regions support the answer",
  "grounding_boxes": [
    {"label": "object or region name", "box": [x1, y1, x2, y2]}
  ],
  "answer": "short final answer"
}

Rules:
1. Keep the reasoning concise and tied to visible evidence in the image.
2. The visual_claim must be a single atomic visual finding that directly supports the answer.
3. The grounding_boxes must localize the region(s) that support the visual_claim.
4. Use tight boxes around the relevant visual evidence whenever possible.
5. Do not ground irrelevant objects, background regions, or the whole image unless the evidence focus requires global scene evidence.
6. If no specific local region is decisive, return grounding_boxes: [].
7. Return raw JSON only.
\end{lstlisting}
\end{tcolorbox}

\paragraph{Grounding boxes and coordinate normalization.}
The \texttt{grounding\_boxes} field localizes the image region(s) that support the visual claim. 
Since different VLMs may emit boxes under different coordinate conventions, we normalize all predicted boxes into the original image pixel coordinate system before evidence diagnosis. 
This shared coordinate frame allows grounding boxes from different agents to be compared consistently.


\subsection{Evidence Diagnosis}
\label{app:evidence-diagnosis}

Evidence Diagnosis evaluates the current round outputs to determine whether the agents have reached evidence-aligned agreement, and otherwise provides diagnostic signals for subsequent Grounded Revision. 
It jointly evaluates two signals: \emph{answer consistency} and \emph{evidence consistency}. 
Answer consistency measures whether agents produce semantically consistent answers, while evidence consistency measures whether their supporting visual evidence is compatible.

\subsubsection{Answer Consistency}

Answer consistency measures whether agents produce semantically consistent answers. 
Let $\tilde{\mathcal{Y}}^{(t)}=\{\tilde{y}_i^{(t)}\}_{i=1}^{M}$ denote the set of semantic-normalized answers produced by $M$ agents at round $t$. 
We categorize answer consistency as:
\begin{equation}
C_y^{(t)} =
\begin{cases}
\textsc{Unanimous}, 
& |\mathrm{unique}(\tilde{\mathcal{Y}}^{(t)})|=1,\\
\textsc{Majority}, 
& \exists \tilde{y}:\mathrm{count}(\tilde{y}) > M/2,\\
\textsc{Divergent}, 
& \text{otherwise}.
\end{cases}
\end{equation}
This signal only considers the normalized answers and does not inspect the visual evidence behind them.

\subsubsection{Evidence Consistency}

Evidence consistency measures whether agents rely on compatible visual evidence. 
We evaluate it using two complementary signals: \emph{spatial alignment} between grounding boxes and \emph{claim alignment} between atomic visual claims. 
An agent pair is considered visually aligned only when both signals are satisfied, and a round is considered evidence-aligned only when all agent pairs are visually aligned.

\paragraph{Spatial alignment.}
Since supporting evidence may contain multiple regions, we compare grounding-box sets via mutual coverage. 
For an agent pair $(i,j)$, the directional coverage from $i$ to $j$ is defined as:
\begin{equation}
\mathrm{Cov}_{i\rightarrow j}^{(t)}
=
\frac{1}{|b_i^{(t)}|}
\sum_{b\in b_i^{(t)}}
\mathbb{I}
\left[
\max_{b'\in b_j^{(t)}} \mathrm{IoU}(b,b') > \tau_{\mathrm{iou}}
\right],
\end{equation}
where $\tau_{\mathrm{iou}}$ is a predefined IoU threshold.

The two agents are spatially aligned only when their box sets mutually cover each other:
\begin{equation}
\mathrm{Align}^{\mathrm{sp},(t)}_{ij}
=
\mathbb{I}
\left[
\mathrm{Cov}_{i\rightarrow j}^{(t)}=1
\land
\mathrm{Cov}_{j\rightarrow i}^{(t)}=1
\right].
\end{equation}
This mutual-coverage criterion avoids treating partial overlap as full evidence alignment when agents share only one region but disagree on other supporting regions.

\paragraph{Claim alignment.}
Spatial overlap alone does not guarantee that agents interpret the evidence consistently. 
Two agents may look at overlapping regions but make incompatible claims about what the regions show. 
We therefore use an LLM judge to compare their atomic visual claims:
\begin{equation}
\mathrm{Align}^{\mathrm{cl},(t)}_{ij}
=
\mathrm{Judge}\big(q,c_i^{(t)},c_j^{(t)}\big),
\end{equation}
where $\mathrm{Align}^{\mathrm{cl},(t)}_{ij}=1$ indicates compatible claims, and $0$ indicates incompatible claims. 
In our implementation, the judge is instantiated with LLaMA-3.2-1B-Instruct. 
The judge only checks whether the two claims describe compatible visual evidence for answering the given question.

\begin{tcolorbox}[
    colback=blue!6,
    colframe=blue!65!black,
    colbacktitle=blue!70!black,
    coltitle=white,
    title=\textbf{Claim Alignment Judge Prompt},
    fonttitle=\bfseries,
    boxrule=0.6pt,
    arc=3pt,
    left=8pt,
    right=8pt,
    top=7pt,
    bottom=7pt,
    breakable
]
\begin{lstlisting}[
    basicstyle=\ttfamily\small,
    columns=fullflexible,
    breaklines=true,
    breakatwhitespace=true,
    keepspaces=true,
    showstringspaces=false,
    frame=none
]
You are a strict evaluator of whether two visual claims refer to compatible evidence for an image question.

Do not judge whether the final answer is correct.
Only judge whether the two claims are grounded in compatible visual evidence for answering the question.

QUESTION:
{question}

CLAIM A:
{visual_claim_i}

CLAIM B:
{visual_claim_j}

Question:
Do Claim A and Claim B refer to compatible visual evidence for answering the question?

Return one compact JSON object only:
{
  "claim_aligned": true or false,
  "reason": "brief explanation"
}
\end{lstlisting}
\end{tcolorbox}

We convert the judge output into a binary alignment value:
\begin{equation}
\mathrm{Align}^{\mathrm{cl},(t)}_{ij}
=
\begin{cases}
1, & \text{if } \texttt{claim\_aligned}=\texttt{true},\\
0, & \text{otherwise}.
\end{cases}
\end{equation}

\paragraph{Evidence alignment.}
An agent pair is considered visually aligned only when both spatial and claim alignment hold:
\begin{equation}
\mathrm{Align}_{ij}^{(t)}
=
\mathbb{I}
\left[
\mathrm{Align}^{\mathrm{sp},(t)}_{ij}=1
\land
\mathrm{Align}^{\mathrm{cl},(t)}_{ij}=1
\right].
\end{equation}

Let $\mathcal{P}=\{(i,j)\mid 1\leq i<j\leq M\}$ be all agent pairs. 
We summarize pairwise alignment into round-level evidence consistency:
\begin{equation}
C_g^{(t)} =
\begin{cases}
\textsc{Aligned}, 
& \forall (i,j)\in\mathcal{P},~\mathrm{Align}_{ij}^{(t)}=1,\\
\textsc{Partial}, 
& \exists (i,j)\in\mathcal{P},~\mathrm{Align}_{ij}^{(t)}=1,\\
\textsc{Dispersed}, 
& \text{otherwise}.
\end{cases}
\end{equation}

\subsubsection{Evidence-Gated Early Exit}

Evidence-Gated Early Exit accepts the current round only when agents achieve both unanimous answer agreement and aligned visual evidence:
\begin{equation}
C_y^{(t)} = \textsc{Unanimous}
\;\land\;
C_g^{(t)} = \textsc{Aligned}.
\end{equation}

If this condition is satisfied, \method{} directly returns the common answer. 
Otherwise, if the maximum number of revision rounds has not been reached, \method{} proceeds to the next Grounded Revision round. 
The revision is guided by the diagnostic summary:
\begin{equation}
D^{(t)}
=
\left\{
C_y^{(t)},
C_g^{(t)},
\mathrm{Align}_{ij}^{(t)}
\right\},
\end{equation}
which provides explicit signals of answer-level and evidence-level consistency. 
If the maximum number of rounds has been reached, the process proceeds to Evidence-Guided Arbitration.

\subsection{Grounded Revision}
\label{app:grounded-revision}

When agents fail to reach evidence-aligned consensus, \method{} treats the current state as uncertain and triggers Grounded Revision. 
The goal is to encourage agents to verify visual evidence before updating their predictions, rather than directly adopting peer rationales.

At round $t>1$, each agent uses the previous-round state as reference, including self outputs, peer outputs, and diagnostic signals. 
Each agent $a_i$ produces:
\begin{equation}
\label{eq:app-grounded-revision}
o_i^{(t)}
=
a_i\left(I, q, f, \mathcal{S}^{(t-1)}\right)
=
\left(
y_i^{(t)}, r_i^{(t)}, b_i^{(t)}, c_i^{(t)}
\right).
\end{equation}
Here, $I$ is the image, $q$ is the question, $f$ is the routed evidence focus, and 
\(\mathcal{S}^{(t-1)}=\{\mathcal{O}^{(t-1)}, \mathcal{D}^{(t-1)}\}\) contains the previous agent outputs and diagnosis results. 
The output keeps the same structure as Eq.~\ref{eq:grounded-answer}: $y_i^{(t)}$ is the answer, $r_i^{(t)}$ is the rationale, $b_i^{(t)}$ denotes the grounding boxes, and $c_i^{(t)}$ is an atomic visual claim explaining how the grounded regions support the answer.

For each agent, the previous-round state is converted into three revision signals. 
First, the \textit{self hypothesis} preserves the agent's own previous answer, rationale, visual claim, and grounding boxes. 
Second, the \textit{peer hypotheses} expose alternative grounded predictions from other agents. 
Third, the \textit{diagnosis summary} provides global and pairwise answer/evidence consistency signals. 
These signals are used as visual references for re-examining the original image, not as authority to be directly followed.

The core revision prompt is shown below.

\begin{tcolorbox}[
    colback=blue!6,
    colframe=blue!65!black,
    colbacktitle=blue!70!black,
    coltitle=white,
    title=\textbf{Grounded Revision Prompt},
    fonttitle=\bfseries,
    boxrule=0.6pt,
    arc=3pt,
    left=8pt,
    right=8pt,
    top=7pt,
    bottom=7pt,
    breakable
]
\begin{lstlisting}[
    basicstyle=\ttfamily\small,
    columns=fullflexible,
    breaklines=true,
    breakatwhitespace=true,
    keepspaces=true,
    showstringspaces=false,
    frame=none
]
You are a visual question-answering agent in a grounded revision round.
The image is the only source of visual evidence.

QUESTION:
{question}

EVIDENCE FOCUS:
{mode}
{additional_reading_instruction}

DIAGNOSIS SUMMARY:
Global:
- answer_consistency: {answer_kind}
- evidence_consistency: {evidence_kind}

Pairwise:
{pairwise_diagnosis}
# Each pair contains:
- evidence_consistency: {evidence_kind}

SELF HYPOTHESIS:
- answer: {self_answer}
- rationale: {self_reasoning}
- visual_claim: {self_visual_claim}
- grounding_boxes: {self_grounding_boxes}

PEER HYPOTHESES:
{peer_hypotheses}
# Each peer contains:
- answer: {peer_answer}
- rationale: {peer_reasoning}
- visual_claim: {peer_visual_claim}
- grounding_boxes: {peer_grounding_boxes}

Decision policy:
1. Re-read the original image independently.
2. Use self and peer hypotheses as visual references, not as authority.
3. Keep your previous answer if the image still supports it.
4. Revise only if a newly verified visual observation better supports another answer.
5. Keep the reasoning concise and grounded in visible evidence.

Output JSON:
{
  "answer": "short final answer",
  "reasoning": "brief image-grounded reasoning",
  "grounding_boxes": [
    {"label": "object name", "box": [x1, y1, x2, y2]}
  ],
  "visual_claim": "one atomic visual finding that directly supports the answer"
}
\end{lstlisting}
\end{tcolorbox}

\subsection{Evidence-Guided Arbitration}
\label{app:evidence-guided-arbitration}

When no evidence-aligned consensus is reached within the maximum number of rounds, \method{} applies Evidence-Guided Arbitration. 
The goal is to preserve vote-based aggregation when it is visually supported, while avoiding the failure mode where a majority answer is selected despite inconsistent visual evidence.

Let the final-round output of agent $i$ follow the structured prediction format:
\begin{equation}
o_i^{(T)}
=
\left(
y_i^{(T)}, r_i^{(T)}, b_i^{(T)}, c_i^{(T)}
\right),
\end{equation}
where $y_i^{(T)}$ is the final-round answer, $r_i^{(T)}$ is the rationale, $b_i^{(T)}$ denotes the grounding boxes, and $c_i^{(T)}$ is the atomic visual claim.

For each candidate answer $y$, we define its supporting agent group as:
\begin{equation}
\mathcal{G}(y)=\{i \mid y_i^{(T)}=y\}.
\end{equation}
Candidate answers are checked in descending order of vote count $|\mathcal{G}(y)|$.

For each candidate with multiple supporting agents, i.e., $|\mathcal{G}(y)|>1$, we compute the internal evidence consistency of its supporting group. 
Let
\begin{equation}
\mathcal{P}_y=\{(i,j)\mid i,j\in\mathcal{G}(y),\, i<j\}
\end{equation}
denote all agent pairs supporting answer $y$. 
Using the pairwise evidence-alignment indicator $\mathrm{Align}_{ij}^{(T)}$ defined in Appendix~\ref{app:evidence-diagnosis}, we define:
\begin{equation}
C_g(\mathcal{G}(y)) =
\begin{cases}
\textsc{Aligned}, 
& \forall (i,j)\in\mathcal{P}_y,~\mathrm{Align}_{ij}^{(T)}=1,\\
\textsc{Partial}, 
& \exists (i,j)\in\mathcal{P}_y,~\mathrm{Align}_{ij}^{(T)}=1,\\
\textsc{Dispersed}, 
& \text{otherwise}.
\end{cases}
\end{equation}

The arbitration rule first selects the highest-voted candidate whose supporting group is evidence-aligned:
\begin{equation}
\hat{y}
=
\operatorname*{first}_{y}
\left[
|\mathcal{G}(y)|>1
\land
C_g(\mathcal{G}(y))=\textsc{Aligned}
\right],
\end{equation}
where candidates are considered in descending order of vote count. 
Thus, vote-based aggregation is accepted when it is supported by aligned visual evidence, while higher-vote candidates with inconsistent evidence can be rejected in favor of visually reliable alternatives.

If no evidence-aligned answer group exists, \method{} falls back to the highest-voted answer. 
Let
\begin{equation}
\mathcal{Y}_{\mathrm{top}}
=
\left\{
y \mid |\mathcal{G}(y)| =
\max_{y'} |\mathcal{G}(y')|
\right\}
\end{equation}
denote the set of top-voted candidate answers. 
If there is a unique top-voted candidate, \method{} returns it:
\begin{equation}
\hat{y}=y,
\quad
y\in\mathcal{Y}_{\mathrm{top}}.
\end{equation}

When multiple candidates are tied, the tie is broken by the strongest grounded single-agent prediction. 
Specifically, let $s_i$ denote a fixed single-agent strength score or ranking for agent $i$, estimated from its individual validation performance, and define the grounded supporting agents for answer $y$ as:
\begin{equation}
\mathcal{G}_{\mathrm{gr}}(y)
=
\left\{
i \in \mathcal{G}(y)
\mid b_i^{(T)} \neq \emptyset
\right\}.
\end{equation}
The final answer is selected as:
\begin{equation}
\hat{y}
=
\operatorname*{arg\,max}_{y\in\mathcal{Y}_{\mathrm{top}}}
\max_{i\in\mathcal{G}_{\mathrm{gr}}(y)} s_i .
\end{equation}
In the rare case where none of the tied candidates has grounding boxes, \method{} applies the same fixed single-agent strength ranking as a deterministic fallback.

Overall, Evidence-Guided Arbitration remains vote-driven in candidate ordering, but uses evidence consistency as a reliability filter before making the final decision. 
When no evidence-aligned group exists, it falls back to the highest-voted answer, with ties resolved by the strongest grounded single-agent prediction. 
This reflects the core intuition that explicit visual support is preferable to unsupported or evidence-inconsistent agreement.

\section{Main Experimental Setup}
\label{app:exp}

\subsection{Benchmark Details}
\label{app:benchmark-details}

We provide the data sources and evaluation focus of each benchmark below. 
All selected benchmarks are closely related to visual grounding: they require models to locate, compare, or reason over specific visual evidence rather than relying solely on simple image-level recognition.

\begin{itemize}
    \item \href{https://huggingface.co/datasets/russwang/vstarbench}{\textbf{V*Bench}}~\citep{wu2024v} evaluates guided visual search and fine-grained localization in high-resolution or visually crowded images.

    \item \href{https://huggingface.co/datasets/Mayfull/whats_up_vlms}{\textbf{WhatsUp}}~\citep{kamath2023s} focuses on grounded spatial-relation understanding, such as distinguishing relative positions and spatial prepositions. Considering the evaluation cost, we randomly sample 1K examples for testing.

    \item \href{https://huggingface.co/datasets/lmms-lab/MME}{\textbf{MME}}~\citep{fu2026mme} covers multimodal perception and cognition tasks, many of which require grounding predictions in specific visual regions or attributes. Considering the evaluation cost, we randomly sample 1K examples for testing.

    \item \href{https://huggingface.co/datasets/lmms-lab/GQA}{\textbf{GQA}}~\citep{hudson2019gqa} evaluates compositional visual reasoning over scene-graph-based questions, requiring models to ground answers in objects, attributes, and relations. Considering the evaluation cost, we randomly sample 1K examples for testing.

    \item \href{https://huggingface.co/datasets/Moenupa/MC-Bench}{\textbf{MC-Bench}}~\citep{xu2025mc} evaluates multi-context visual grounding, where models localize target instances across image pairs based on open-ended text prompts. We adapt it into a VQA-style evaluation to test whether models can identify the visually grounded target across images. Considering the evaluation cost, we randomly sample 1K examples for testing.
    
    \item \href{https://huggingface.co/datasets/MMVP/MMVP}{\textbf{MMVP}}~\citep{tong2024eyes} tests fine-grained visual patterns using CLIP-blind image pairs that require localized visual discrimination.
\end{itemize}

\subsection{Baselines Details}
\label{app:baselines}

We compare \method{} with representative single-agent and multi-agent inference-time baselines. 
All baselines are evaluated on the same benchmark inputs and under the same answer-normalization and evaluation protocol. 
For single-agent methods, we use the same visual question answering format as \method{} but without cross-agent communication. 
For multi-agent methods, we use the same three-agent committee as \method{}, consisting of Qwen3-VL, GLM-4.6V, and InternVL3.5.

\paragraph{Zero-shot CoT}~\cite{wei2022chain,kojima2022large}.
Zero-shot Chain-of-Thought prompts a single agent to reason step by step before producing the final answer. 
Given an image and a question, the agent generates evidence-grounded reasoning and then outputs a compact final answer. 
We evaluate Zero-shot CoT separately for each individual agent in the committee to characterize the base capability of each VLM.

\begin{tcolorbox}[
    colback=blue!6,
    colframe=blue!65!black,
    colbacktitle=blue!70!black,
    coltitle=white,
    title=\textbf{Zero-shot CoT Prompt},
    fonttitle=\bfseries,
    boxrule=0.6pt,
    arc=3pt,
    left=8pt,
    right=8pt,
    top=7pt,
    bottom=7pt,
    breakable
]
\begin{lstlisting}[
    basicstyle=\ttfamily\small,
    columns=fullflexible,
    breaklines=true,
    breakatwhitespace=true,
    keepspaces=true,
    showstringspaces=false,
    frame=none
]
Analyze the image question carefully and reason step by step from the visual evidence before deciding on the final answer.

Question: {question}

Output schema:
{
  "reasoning": "step-by-step evidence-grounded chain of thought",
  "answer": "short final answer"
}

Rules:
1. Use explicit multi-step reasoning grounded in the image and question.
2. Keep the reasoning focused and concrete rather than verbose.
3. Return raw JSON only.
\end{lstlisting}
\end{tcolorbox}

\paragraph{Self-Consistency}~\cite{wang2022self}.
Self-Consistency samples multiple reasoning paths from a single model and aggregates their final answers by voting. 
For each question, we query the same backbone multiple times with the Zero-shot CoT prompt above. 
Each response contains its own reasoning path and final answer. 
We then discard the intermediate reasoning and apply majority voting over the normalized final answers. 
In our experiments, we use Qwen3-VL as the backbone for Self-Consistency, since it is the strongest individual agent among the three VLMs.

\begin{tcolorbox}[
    colback=blue!6,
    colframe=blue!65!black,
    colbacktitle=blue!70!black,
    coltitle=white,
    title=\textbf{Self-Consistency Prompt},
    fonttitle=\bfseries,
    boxrule=0.6pt,
    arc=3pt,
    left=8pt,
    right=8pt,
    top=7pt,
    bottom=7pt,
    breakable
]
\begin{lstlisting}[
    basicstyle=\ttfamily\small,
    columns=fullflexible,
    breaklines=true,
    breakatwhitespace=true,
    keepspaces=true,
    showstringspaces=false,
    frame=none
]
We use the same prompt as Zero-shot CoT and sample multiple independent responses.

Aggregation:
1. Extract the final answer from each sampled response.
2. Select the final answer by majority voting.
\end{lstlisting}
\end{tcolorbox}

\paragraph{Self-Refine}~\cite{madaan2023self}.
Self-Refine iteratively improves a model's own answer using self-generated feedback. 
For each sample, the model first generates an initial answer with the Zero-shot CoT prompt. 
It then critiques its own reasoning and answer, and finally produces a refined response conditioned on the critique. 
The final refined answer is used for evaluation. 
As with Self-Consistency, we instantiate Self-Refine with Qwen3-VL.










We use the Zero-shot CoT prompt to obtain the initial answer, and then apply Self-Refine with the following critique-and-revision prompts.

\begin{tcolorbox}[
    colback=blue!6,
    colframe=blue!65!black,
    colbacktitle=blue!70!black,
    coltitle=white,
    title=\textbf{Self-Refine Prompt},
    fonttitle=\bfseries,
    boxrule=0.6pt,
    arc=3pt,
    left=8pt,
    right=8pt,
    top=7pt,
    bottom=7pt,
    breakable
]
\begin{lstlisting}[
    basicstyle=\ttfamily\small,
    columns=fullflexible,
    breaklines=true,
    breakatwhitespace=true,
    keepspaces=true,
    showstringspaces=false,
    frame=none
]
You are refining an answer to a visual question.
First critique whether the current reasoning and answer are supported by the image.
Then revise the answer only if the critique identifies a real visual error.

Input:
- question: {question}
- current_reasoning: {reasoning}
- current_answer: {answer}

Output JSON:
{
  "critique": "brief critique of the main potential issue, grounded in the image",
  "revised_reasoning": "brief revised reasoning grounded in the image",
  "revised_answer": "short final answer"
}

Rules:
1. Focus the critique on the most important possible error in the reasoning or answer.
2. If the current answer is still supported by the image, keep it unchanged.
3. Revise the answer only when the image provides evidence for the change.
4. Do not introduce information that is not visible in the image.
5. Return raw JSON only.
\end{lstlisting}
\end{tcolorbox}

\paragraph{Multi-Agent Debate}~\cite{du2024improving,liang2024encouraging}.
Multi-Agent Debate lets multiple agents exchange their answers and textual rationales over multiple rounds. 
In the first round, each agent independently answers the question using the Zero-shot CoT prompt. 
In later rounds, each agent observes the other agents' previous answers and rationales, then revises its own prediction. 
Unlike \method{}, this baseline exchanges only textual rationales and final answers; it does not explicitly compare grounded visual evidence. 
We evaluate two aggregation variants:

\begin{itemize}
    \item \textit{Majority-vote aggregation.} 
    The final answer is selected by voting over the agents' last-round predictions. 
    If multiple answers receive the same number of votes, we use a deterministic tie-breaking rule based on the implementation order of the agents.

    \item \textit{Judge-based aggregation.} 
    The agents first perform the same multi-round textual debate. 
    Then, a separate judge model receives the question, candidate answers, and debate traces, and selects the final answer.
\end{itemize}

\begin{tcolorbox}[
    colback=blue!6,
    colframe=blue!65!black,
    colbacktitle=blue!70!black,
    coltitle=white,
    title=\textbf{Multi-Agent Debate Prompt},
    fonttitle=\bfseries,
    boxrule=0.6pt,
    arc=3pt,
    left=8pt,
    right=8pt,
    top=7pt,
    bottom=7pt,
    breakable
]
\begin{lstlisting}[
    basicstyle=\ttfamily\small,
    columns=fullflexible,
    breaklines=true,
    breakatwhitespace=true,
    keepspaces=true,
    showstringspaces=false,
    frame=none
]
These are the solutions to the problem from other agents.

Question: {question}
Previous response: {previous_text}
Other agent solutions: {peer_json}

Based on the opinions of the other agents, give an updated response.

Output schema:
{
  "reasoning": "step-by-step evidence-grounded reasoning",
  "answer": "short final answer"
}

Rules:
1. Consider peers, but do not follow them blindly.
2. Explain step by step how the peer evidence changes or confirms your view.
3. Keep the reasoning concrete and tied to the image question.
4. Return raw JSON only.
\end{lstlisting}
\end{tcolorbox}

\begin{tcolorbox}[
    colback=blue!6,
    colframe=blue!65!black,
    colbacktitle=blue!70!black,
    coltitle=white,
    title=\textbf{Debate Judge Prompt},
    fonttitle=\bfseries,
    boxrule=0.6pt,
    arc=3pt,
    left=8pt,
    right=8pt,
    top=7pt,
    bottom=7pt,
    breakable
]
\begin{lstlisting}[
    basicstyle=\ttfamily\small,
    columns=fullflexible,
    breaklines=true,
    breakatwhitespace=true,
    keepspaces=true,
    showstringspaces=false,
    frame=none
]
You are given an image question and the full state of a multi-round debate among several vision-language agents.

Question: {question}

Debate states:
{debate_text}

Task:
1. Read the image yourself.
2. Use the debate states only as auxiliary evidence.
3. Identify all candidate answers that appeared in the debate states.
4. Select the single best final answer from these candidate answers only.

Output schema:
{
  "reasoning": "brief image-grounded adjudication that explains why the selected candidate is best",
  "answer": "one candidate answer copied from the debate states"
}

Rules:
1. The image is the source of truth; do not blindly follow the debaters.
2. You must choose one answer that already appears in the debate states.
3. If multiple candidates are plausible, choose the one best supported by the image.
4. Return raw JSON only.
\end{lstlisting}
\end{tcolorbox}

\paragraph{ReConcile}~\cite{chen2024reconcile}.
ReConcile is a confidence-driven multi-agent discussion framework. 
Each agent first provides an answer with a confidence score. 
Then, agents review grouped peer answers, justifications, and confidences before updating their predictions. 
After the final discussion round, we group semantically equivalent answers and select the answer group with the highest aggregated confidence.

\begin{tcolorbox}[
    colback=blue!6,
    colframe=blue!65!black,
    colbacktitle=blue!70!black,
    coltitle=white,
    title=\textbf{ReConcile Prompts and Aggregation Rule},
    fonttitle=\bfseries,
    boxrule=0.6pt,
    arc=3pt,
    left=8pt,
    right=8pt,
    top=7pt,
    bottom=7pt,
    breakable
]
\begin{lstlisting}[
    basicstyle=\ttfamily\small,
    columns=fullflexible,
    breaklines=true,
    breakatwhitespace=true,
    keepspaces=true,
    showstringspaces=false,
    frame=none
]
[Initial answer with confidence]
Analyze the image question carefully, reason from the visual evidence, then provide your final answer and confidence.

Question: {question}

Output JSON:
{
  "reasoning": "brief evidence-grounded reasoning",
  "answer": "short final answer",
  "confidence": 0.0
}

Rules:
1. Base your answer on the image and question.
2. Confidence must be a number between 0 and 1.
3. Keep the reasoning focused and concrete.
4. Return raw JSON only.

[Reconcile]
You are in a round-table conference with other agents.
Review grouped peer answers, justifications, and confidences, then update your answer and confidence.

Question: {question}

Previous response:
{previous_text}

Grouped peer views:
{peer_json}

Output JSON:
{
  "reasoning": "brief evidence-grounded reasoning after reviewing peer views",
  "answer": "short final answer",
  "confidence": 0.0
}

Rules:
1. Review each answer group and compare the supporting justifications.
2. Keep your answer if it remains best supported by the image.
3. Change your answer only if another group provides more convincing visual evidence.
4. Confidence must reflect your final belief after reviewing all groups.
5. Return raw JSON only.

[Final confidence-aware aggregation]
After the last discussion round, group semantically equivalent final answers.
For each answer group y, compute its aggregation score as the sum of confidences from agents supporting y:

score(y) = sum(confidence_i for agents whose final answer is y)

Select the answer group with the highest score as the final prediction.
If there is a tie, choose the group with more supporting agents.
If the tie still remains, choose the group with the higher average confidence.
\end{lstlisting}
\end{tcolorbox}

\paragraph{DART}~\cite{sivakumaran2026dart}.
DART is a tool-augmented multi-agent framework that recruits external visual tools based on inter-agent disagreement. 
It first collects initial answers from multiple agents, then uses a recruitment module to decide which tools are needed to resolve the disagreement. 
The recruited tool outputs are injected into a subsequent discussion round, together with tool-aligned agreement scores that measure how well each agent prediction is supported by the tool evidence. 
A final aggregator then selects the answer based on post-discussion agent outputs, tool outputs, and agreement scores.

Following the original DART setup, we use a heterogeneous pool of expert visual tools. 
Specifically, the tool pool includes \textit{grounding}, \textit{object detection}, \textit{OCR}, \textit{spatial reasoning}, \textit{captioning}, \textit{attribute detection}, and \textit{reasoning}. 
We instantiate these tools following DART: GroundingDINO is used for grounding~\citep{liu2024grounding}, YOLOv11 for object detection~\citep{khanam2024yolov11}, SpaceLLaVA for spatial reasoning~\citep{chen2024spatialvlm}, OCR-Qwen for OCR, and InternVL-2.5 MPO for captioning, attribute detection, and reasoning~\citep{wang2024enhancing}. 
This follows the design principle of DART: specialized tools provide additional perceptual evidence for resolving inter-agent disagreement, while the VLM-based tools are executed by an independent model to reduce overlap between answering agents and tool-side information.

\begin{tcolorbox}[
    colback=blue!6,
    colframe=blue!65!black,
    colbacktitle=blue!70!black,
    coltitle=white,
    title=\textbf{DART Prompts and Tool Design},
    fonttitle=\bfseries,
    boxrule=0.6pt,
    arc=3pt,
    left=8pt,
    right=8pt,
    top=7pt,
    bottom=7pt,
    breakable
]
\begin{lstlisting}[
    basicstyle=\ttfamily\small,
    columns=fullflexible,
    breaklines=true,
    breakatwhitespace=true,
    keepspaces=true,
    showstringspaces=false,
    frame=none
]
[Initial answer]
Use the same initial answer-with-confidence prompt as ReConcile.

[Tool recruitment]
You are a recruitment agent for multimodal reasoning.
Inspect the disagreement among agents and decide which visual tools are needed.

Question:
{question}

Candidate solutions:
{solutions_json}

Available tools:
- grounding: localize visual regions relevant to the question or candidate answers.
- object_detection: detect objects mentioned in the question or agent solutions.
- ocr: read visible text, letters, numbers, labels, or symbols from the image.
- spatial_reasoning: analyze spatial relations, relative positions, distances, or cross-region comparisons.
- captioning: summarize global or scene-level visual content relevant to the question.
- attribute_detection: identify visual attributes such as color, shape, material, state, or markings.
- reasoning: provide additional visual reasoning over the image and candidate answers.

Output JSON:
{
  "selected_tools": ["tool_name_1", "tool_name_2"],
  "tool_queries": [
    {
      "tool_name": "tool_name",
      "query": "specific query or input for this tool"
    }
  ],
  "reasoning": "brief explanation of which disagreement each selected tool resolves"
}

Rules:
1. Select only tools that are useful for resolving the disagreement.
2. Select "grounding" when agents disagree about where the relevant evidence is located.
3. Select "object_detection" when agents disagree about the presence or identity of objects.
4. Select "ocr" when the question depends on visible text, letters, numbers, labels, or symbols.
5. Select "spatial_reasoning" when agents disagree about relative positions, directions, distances, or spatial configurations.
6. Select "captioning" when global scene context may resolve the disagreement.
7. Select "attribute_detection" when agents disagree about visual attributes such as color, shape, material, state, or markings.
8. Select "reasoning" when the disagreement requires additional visual reasoning beyond direct perception.
9. Return raw JSON only.

[Expert tool execution]
Each selected tool is executed with its corresponding query.
Tool implementations:
- grounding: GroundingDINO.
- object_detection: YOLOv11.
- spatial_reasoning: SpaceLLaVA.
- ocr: OCR-Qwen.
- captioning / attribute_detection / reasoning: InternVL-2.5 MPO.

Tool output format:
{
  "tool_name": "tool_name",
  "query": "tool query",
  "tool_output": "concise tool result relevant to the question",
  "evidence_summary": "brief summary of the visual evidence found by the tool",
  "grounding_boxes": [
    {"label": "relevant visual region if applicable", "box": [x1, y1, x2, y2]}
  ]
}

[Tool agreement scoring]
Given the tool outputs, score how well each agent prediction is supported by the tool evidence.

Question:
{question}

Agent solutions:
{solutions_json}

Tool outputs:
{tool_json}

Output JSON:
{
  "agreement_scores": [
    {
      "agent": "agent name",
      "score": 0.0,
      "reason": "brief explanation of agreement or disagreement with the tool evidence"
    }
  ]
}

Rules:
1. Score each agent between 0 and 1.
2. A high score means the agent's answer and reasoning are supported by the tool outputs.
3. A low score means the agent's answer conflicts with or is unsupported by the tool outputs.
4. Use the tool outputs as auxiliary evidence, not as the only criterion.
5. Return raw JSON only.

[Tool-assisted discussion]
You are in a tool-assisted multi-agent discussion.
Review the grouped agent solutions, tool outputs, and tool-agreement scores, then update your answer.

Question:
{question}

Previous response:
{previous_text}

Grouped agent solutions:
{grouped_json}

Tool outputs:
{tool_json}

Agreement scores:
{score_json}

Output JSON:
{
  "reasoning": "brief evidence-grounded reasoning after reviewing the tool evidence",
  "answer": "short final answer",
  "confidence": 0.0
}

Rules:
1. Prefer answers supported by reliable tool outputs, while keeping the original image question central.
2. Use agreement scores as auxiliary evidence, not as the only criterion.
3. Keep your answer if it remains best supported by the image and tool evidence.
4. Change your answer only when another candidate is better supported by visual evidence.
5. Confidence must be a number between 0 and 1.
6. Return raw JSON only.

[Final aggregator]
Choose the best final answer after reviewing post-discussion agent solutions, tool outputs, and tool-agreement scores.

Question:
{question}

Post-discussion solutions:
{discussion_json}

Tool outputs:
{tools_json}

Agreement scores:
{scores_json}

Candidate answers:
{candidate_answers}

Output JSON:
{
  "reasoning": "brief explanation of why the selected candidate is best supported",
  "answer": "one candidate answer copied from Candidate answers"
}

Rules:
1. Select exactly one answer from Candidate answers.
2. Do not invent a new answer or output an answer not proposed by any agent.
3. Prefer answers supported by reliable tool outputs.
4. Use vote counts, confidence scores, and tool-agreement scores together.
5. Do not rely on tool scores alone if image-grounded reasoning contradicts them.
6. Return raw JSON only.
\end{lstlisting}
\end{tcolorbox}

\subsection{Implementation Details}
\label{app:evaluations}

We implement all VLM inference with vLLM~\citep{kwon2023efficient}. 
Unless otherwise specified, models are decoded greedily with temperature $0$ and \texttt{max\_tokens=4096}. 
For Self-Consistency, we sample three reasoning paths with temperature $0.8$. 
For \method{}, we set the maximum number of inference rounds to 2 and use $\tau_{\mathrm{iou}}=0.4$ in Evidence Diagnosis, following the best-performing settings in the parameter ablations (Section~\ref{sec:parameter-ablations}). 

For the baselines, Self-Consistency samples three reasoning paths, Self-Refine performs three refinement rounds, and multi-agent baselines follow a standard three-round discussion protocol. 
For debate-based methods, if all agents reach answer agreement before the final round, we terminate the discussion early and return the common answer. 
All experiments are conducted on 8 NVIDIA A100 GPUs.

\section{Additional Case Studies}
\label{app:case-study}

We provide representative case studies to illustrate how \method{} differs from standard multi-agent discussion.
Each case is organized as a complete table containing the image, question, answer options, ground-truth answer, and a round-by-round comparison between \method{} and standard MAD.
Instead of only reporting final answer letters, we summarize how each agent reads the visual evidence, how its answer changes across rounds, and why the final decision is accepted, revised, or arbitrated by \method{}.

We include one representative case for each decision mechanism:
(1) \textit{evidence-aligned early exit}, where all agents agree on both the answer and supporting evidence in the first round;
(2) \textit{grounded revision to aligned consensus}, where the first round is not reliable enough but the second round reaches evidence-aligned agreement;
and (3) \textit{evidence-guided arbitration}, where the second round still has a 2:1 answer split, but the majority answer is supported by an evidence-aligned agent group.


\subsection{Evidence-Aligned Early Exit}
\label{app:case-aligned-consensus}

This case shows that \method{} can avoid unnecessary discussion when the first-round consensus is already visually reliable.


\paragraph{Case 1: Absence-of-evidence OCR.}
Table~\ref{tab:case-mmvp62} shows a keyboard OCR case.
The key difficulty is that the correct answer depends on verifying the absence of a queried letter, rather than locating a positive object.
In \method{}, all three agents inspect the visible keyboard crop and agree that no visible key contains ``J''.
Because both the answer and the negative visual evidence are aligned, \method{} exits immediately.
Standard MAD, however, repeatedly hallucinates a full QWERTY row and converts this layout prior into a wrong majority.

\begin{table*}[t]
\caption{
Case 1. \method{} safely early-exits because all agents agree on the same negative visual evidence, while MAD hallucinates a full keyboard layout.
}
\centering
\scriptsize
\setlength{\tabcolsep}{4pt}
\renewcommand{\arraystretch}{1.08}
\begin{tabular}{p{0.08\textwidth}p{0.43\textwidth}p{0.43\textwidth}}
\toprule
\caseheader
{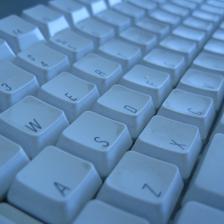}
{Case 1: Evidence-aligned early exit / absence-of-evidence OCR.}
{MMVP, sample 62.}
{Can you see letter ``J'' in the image?}
{A = Yes; B = No.}
{B = No.}
\\
\midrule
\textbf{Stage} & \textbf{\method{} (default R2)} & \textbf{Standard MAD (3 rounds)} \\
\midrule

\rowcolor{casegray}
R1
&
\agentout{GLM}{B = No}
{It examines the visible keyboard keys and lists letters such as W, E, R, S, D, X, A, and Z, concluding that no key labeled ``J'' is visible.}

\agentout{Qwen}{B = No}
{It checks the shown crop rather than assuming a complete keyboard layout; it notes that the region around H and K is not visible and that ``J'' is absent from the visible keys.}

\agentout{InternVL}{B = No}
{It also restricts the judgment to the visible keys and states that the letter ``J'' is not shown.}

\caseDiagnosis{answer = \textsc{Unanimous}; evidence = \textsc{Aligned}. The three agents share the same negative visual claim.}
&
\agentout{GLM}{A = Yes}
{It lists a full QWERTY keyboard, including H, J, and K, and claims that ``J'' is clearly visible.}

\agentout{Qwen}{A = Yes}
{It similarly assumes a complete middle row and says ``J'' is visible between H and K.}

\agentout{InternVL}{B = No}
{It correctly checks only the visible crop and says that ``J'' is not present.}
\\

\rowcolor{caseblue}
R2
&
\textbf{Not used.}
\method{} exits at R1 because the unanimous answer is supported by aligned negative evidence.
&
\agentout{GLM}{B = No}
{It temporarily corrects itself after re-checking the actual visible keys and finding no clear ``J''.}

\agentout{Qwen}{A = Yes}
{It maintains the full-keyboard-row hallucination and treats the ``No'' answer as a misinterpretation.}

\agentout{InternVL}{A = Yes}
{It is pulled toward the peer hallucination and repeats that ``J'' is visible between H and K.}
\\

\rowcolor{casegray}
R3
&
\textbf{Not used.}
The prediction has already been accepted by evidence-gated early exit.
&
\agentout{GLM}{A = Yes}
{It follows the peer claim and again says the middle row contains H, J, and K.}

\agentout{Qwen}{A = Yes}
{It claims that ``J'' is legible and even attributes this interpretation to peer evidence.}

\agentout{InternVL}{B = No}
{It remains grounded in the actual crop and says the ``J'' key is not visible.}
\\

\rowcolor{casegreen}
Final
&
\caseFinal{B = No}
{All agents agree that no visible key contains ``J''.}
&
\cellcolor{casered}
\caseFailure{A = Yes}
{Text-only discussion amplifies a keyboard-layout prior and turns a hallucinated key into the majority answer.}
\\

\bottomrule
\end{tabular}

\label{tab:case-mmvp62}
\end{table*}


\subsection{Grounded Revision to Aligned Consensus}
\label{app:case-revision}

This case shows how \method{} uses grounded revision when the first-round state is not reliable enough to accept.
The system asks agents to re-check the image with self and peer evidence as references, and accepts the prediction only after answer and evidence become aligned.


\paragraph{Case 2: Spatial relation correction.}
Table~\ref{tab:case-vstar177} shows a left-right relation case.
The question asks whether the soccer ball is on the left or right side of the long bench.
In the first round, GLM attends to an incorrect visual region and selects the wrong relation, while Qwen and InternVL focus on the correct soccer ball--bench pair.
Because the evidence is only partially aligned, \method{} does not immediately accept the majority.
After grounded revision, GLM re-checks the image using peer evidence as reference, corrects its visual focus, and all agents converge to an evidence-aligned consensus.
Standard MAD instead collapses to the wrong relation through text-only discussion.

\begin{table*}[t]
\caption{
Case 2. \method{} uses grounded revision to correct a wrong visual focus and reaches evidence-aligned consensus in the second round.
}
\centering
\scriptsize
\setlength{\tabcolsep}{4pt}
\renewcommand{\arraystretch}{1.08}
\begin{tabular}{p{0.08\textwidth}p{0.43\textwidth}p{0.43\textwidth}}
\toprule
\caseheader
{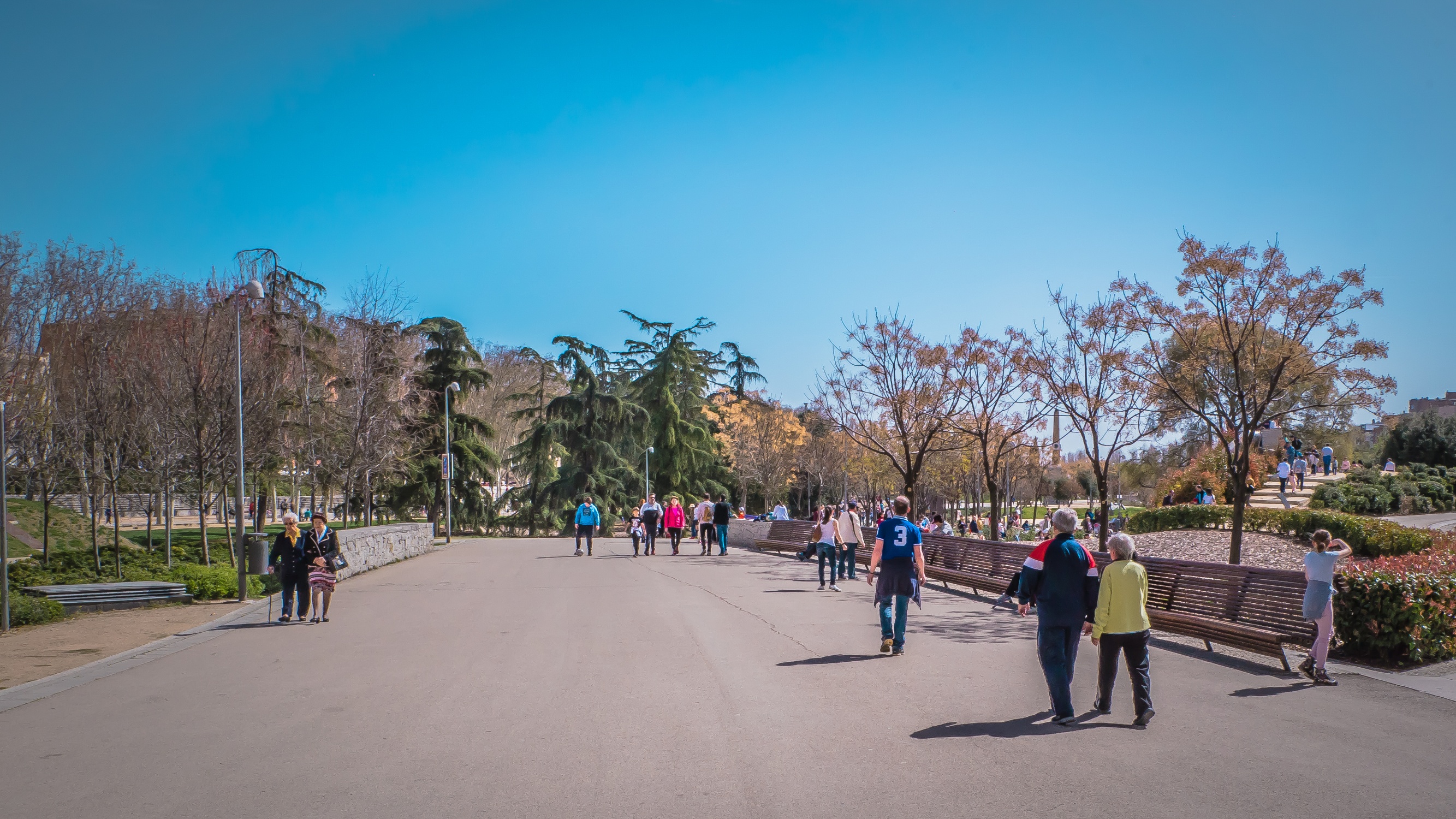}
{Case 2: Grounded revision to aligned consensus / spatial relation.}
{V*Bench, sample 177.}
{Is the soccer ball on the left or right side of the long bench?}
{A = right; B = left.}
{B = left.}
\\
\midrule
\textbf{Stage} & \textbf{\method{} (default R2)} & \textbf{Standard MAD (3 rounds)} \\
\midrule

\rowcolor{casegray}
R1
&
\agentout{GLM}{A = right}
{It focuses on an incorrect region around the bench area and fails to ground the correct soccer ball--bench pair, leading to the wrong right-side relation.}

\agentout{Qwen}{B = left}
{It identifies the long bench and the soccer ball, and grounds the answer in the ball being on the left side of the bench.}

\agentout{InternVL}{B = left}
{It also focuses on the correct soccer ball--bench pair and selects the left-side relation.}

\caseDiagnosis{answer = \textsc{Majority}; evidence = \textsc{Partial}. The B-supporting agents align on the correct ball--bench relation, while GLM relies on a different visual region.}
&
\agentout{GLM}{A = right}
{It selects the opposite relation based on an incorrect visual focus.}

\agentout{Qwen}{A = right}
{It is pulled to the wrong side in the text-only discussion.}

\agentout{InternVL}{B = left}
{It gives the correct relation.}
\\

\rowcolor{caseblue}
R2
&
\agentout{GLM}{B = left}
{After re-checking the image with the peer-highlighted ball--bench evidence, it corrects its visual focus and selects B.}

\agentout{Qwen}{B = left}
{It maintains the same grounded relation: the soccer ball is on the left side of the long bench.}

\agentout{InternVL}{B = left}
{It also preserves the same spatial relation evidence.}

\caseDiagnosis{answer = \textsc{Unanimous}; evidence = \textsc{Aligned}. All agents now support B with the same soccer ball--bench relation.}
&
\agentout{GLM}{A = right}
{It maintains A.}

\agentout{Qwen}{A = right}
{It maintains A.}

\agentout{InternVL}{A = right}
{It switches from the correct answer to the wrong consensus.}
\\

\rowcolor{casegray}
R3
&
\textbf{Not used.}
The round-2 prediction is accepted because both answer and visual evidence are aligned.
&
\textbf{Not used.}
MAD already reaches an incorrect round-2 consensus.
\\

\rowcolor{casegreen}
Final
&
\caseFinal{B = left}
{All agents align on the soccer ball being on the left side of the long bench.}
&
\cellcolor{casered}
\caseFailure{A = right}
{MAD suppresses the initially correct visual evidence and forms a wrong text-level consensus.}
\\

\bottomrule
\end{tabular}

\label{tab:case-vstar177}
\end{table*}
\subsection{Evidence-Guided Arbitration with an Aligned Supporting Group}
\label{app:case-arbitration}

This case shows how \method{} resolves a final 2:1 split.
The method remains vote-driven, but only accepts a candidate when its supporting agents provide aligned visual evidence.
Thus, a majority answer can be selected when it is also evidence-consistent, while visually unreliable majorities are filtered out.


\paragraph{Case 3: Fine-grained spot counting.}
Table~\ref{tab:case-mmvp48} shows a fine-grained counting case.
The question asks for the number of spots on the animal.
The initial majority is wrong because GLM and InternVL count five.
After grounded revision, InternVL aligns with Qwen's six-spot readout, while GLM remains on five.
Although the second round is still not unanimous, the B-supporting majority is evidence-aligned.
\method{} therefore selects B through Evidence-Guided Arbitration.
By contrast, MAD reinforces the wrong initial majority and converges to an incorrect count.

\begin{table*}[t]
\caption{
Case 3. \method{} uses Evidence-Guided Arbitration to select the majority answer supported by an aligned fine-grained counting group.
}
\centering
\scriptsize
\setlength{\tabcolsep}{4pt}
\renewcommand{\arraystretch}{1.08}
\begin{tabular}{p{0.08\textwidth}p{0.43\textwidth}p{0.43\textwidth}}
\toprule
\caseheader
{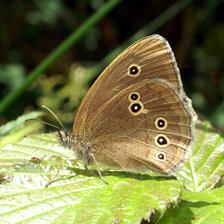}
{Case 3: Evidence-guided arbitration / fine-grained counting.}
{MMVP, sample 48.}
{How many spots are on the animal in the image?}
{A = 5; B = 6.}
{B = 6.}
\\
\midrule
\textbf{Stage} & \textbf{\method{} (default R2)} & \textbf{Standard MAD (3 rounds)} \\
\midrule

\rowcolor{casegray}
R1
&
\agentout{GLM}{A = 5}
{It counts five visible spots on the animal.}

\agentout{Qwen}{B = 6}
{It counts six distinct circular spots, treating the visible spot pattern as the decisive evidence.}

\agentout{InternVL}{A = 5}
{It initially counts five spots.}

\caseDiagnosis{answer = \textsc{Majority}; evidence = \textsc{Partial}. The initial A majority is not accepted because the detailed spot readout is inconsistent.}
&
\agentout{GLM}{A = 5}
{It counts five.}

\agentout{Qwen}{B = 6}
{It counts six.}

\agentout{InternVL}{A = 5}
{It also counts five.}
\\

\rowcolor{caseblue}
R2
&
\agentout{GLM}{A = 5}
{It still counts five and remains dispersed from the B-supporting evidence.}

\agentout{Qwen}{B = 6}
{It maintains the six-spot readout after re-checking the visible pattern.}

\agentout{InternVL}{B = 6}
{It revises after re-checking the image and now counts six distinct spots.}

\caseDiagnosis{answer = \textsc{Majority}; evidence = \textsc{Partial}, with an aligned B-supporting group.}
&
\agentout{GLM}{B = 6}
{It temporarily switches to six.}

\agentout{Qwen}{A = 5}
{It is pulled away from its correct six-spot readout.}

\agentout{InternVL}{A = 5}
{It maintains five.}
\\

\rowcolor{casegray}
R3
&
\textbf{Not used.}
Evidence-Guided Arbitration selects the aligned B-supporting group.
&
\agentout{GLM}{A = 5}
{It switches back to five.}

\agentout{Qwen}{A = 5}
{It follows the wrong five-spot consensus.}

\agentout{InternVL}{A = 5}
{It maintains five.}
\\

\rowcolor{casegreen}
Final
&
\caseFinal{B = 6}
{Qwen and InternVL align on six visible spots.}
&
\cellcolor{casered}
\caseFailure{A = 5}
{MAD reinforces the wrong initial majority and collapses to an incorrect count.}
\\

\bottomrule
\end{tabular}

\label{tab:case-mmvp48}
\end{table*}
\end{document}